\documentclass[letterpaper, 10 pt, journal, twoside]{IEEEtran}

\usepackage{graphicx}
\usepackage{amsmath,amssymb,amsthm,mathabx}
\usepackage{booktabs}
\usepackage{algorithmic}
\usepackage[linesnumbered,ruled,vlined]{algorithm2e}
\usepackage{acronym}
\usepackage{enumitem}
\usepackage{hyperref}
\usepackage{balance}
\usepackage{xspace}
\usepackage[skip=3pt,font=small]{subcaption}
\usepackage[skip=3pt,font=small]{caption}
\usepackage[capitalise]{cleveref}
\usepackage{tabularx,colortbl,multirow,array,makecell}
\usepackage{overpic,wrapfig}
\usepackage{cite}
\usepackage[misc]{ifsym}

\makeatletter
\DeclareRobustCommand\onedot{\futurelet\@let@token\@onedot}
\def\@onedot{\ifx\@let@token.\else.\null\fi\xspace}
\def\eg{\emph{e.g}\onedot} 

\def\ie{\emph{i.e}\onedot}

\def\wrt{w.r.t\onedot} 

\def\etal{\emph{et al}\onedot}
\makeatother

\graphicspath{{figures/}}

\crefname{algocf}{Alg.}{Algs.}
\Crefname{algocf}{Algrithm}{Algrithm}

\acrodef{mala}[MALA]{Metropolis-adjusted Langevin algorithm}
\acrodef{mep}[MEPs]{minimum energy pathways}

\definecolor{myred}{rgb}{0.75, 0.34, 0.34}
\definecolor{myyellow}{rgb}{0.95, 0.76, 0.52}
\definecolor{mygreen}{rgb}{0.33, 0.55, 0.55}

\begin{document}

\title{Synthesizing Diverse and Physically Stable Grasps with Arbitrary Hand Structures\\using Differentiable Force Closure Estimator}

\author{Tengyu Liu$^{1}$, Zeyu Liu$^1$, Ziyuan Jiao$^1$, Yixin Zhu$^{2,3,\dagger}$, and Song-Chun Zhu$^{2,3,4}$%
\thanks{Manuscript received: April, 1$^\mathrm{st}$, 2021; Revised July, 16$^\mathrm{th}$, 2021; Accepted August, 19$^\mathrm{th}$, 2021.}%
\thanks{This paper was recommended for publication by Editor Tamim Asfour upon evaluation of the Associate Editor and Reviewers' comments.}%
\thanks{$\dagger$ Corresponding email: {\tt{y@bigai.ai}}. See additional material on \url{https://sites.google.com/view/ral2021-grasp/}.}
\thanks{$^{1}$ UCLA Center for Vision, Cognition, Learning, and Autonomy (VCLA).}%
\thanks{$^{2}$ Beijing Institute for General Artificial Intelligence (BIGAI).}%
\thanks{$^{3}$ Peking University.\quad{}$^{4}$ Tsinghua University.}%
\thanks{Digital Object Identifier (DOI): see top of this page.}%
}

\markboth{IEEE Robotics and Automation Letters. Preprint Version. Accepted August, 2021}
{Liu \MakeLowercase{\textit{et al.}}: Efficient Synthesis of Diverse and Physically Stable Grasps}

\maketitle

\begin{abstract}
Existing grasp synthesis methods are either analytical or data-driven. The former one is oftentimes limited to specific application scope. The latter one depends heavily on demonstrations~\cite{bohg2013data}, thus suffers from generalization issues; \eg, models trained with human grasp data would be difficult to transfer to 3-finger grippers. To tackle these deficiencies, we formulate a fast and differentiable force closure estimator, capable of producing diverse and physically stable grasps with \emph{arbitrary} hand structures, \emph{without any training data}. Although force closure has commonly served as a measure of grasp quality, it has not been widely adopted as an optimization objective for grasp synthesis primarily due to its high computational complexity; in comparison, the proposed differentiable method can test a force closure within milliseconds. In experiments, we validate the proposed method's efficacy in six different settings.
\end{abstract}


\IEEEpeerreviewmaketitle

\section{Introduction}

\IEEEPARstart{G}{rasp} synthesis has been a challenging task due to the complexity of hand kinematics. Although force closure has been commonly accepted to evaluate the quality of the generated grasps, researchers usually avoid using it as an optimization objective: Computing force closure requires solving for contact forces, which is an optimization problem itself. As a result, using force closure as the optimization objective in grasp synthesis would produce a notoriously slow and nested optimization problem. Instead, researchers have primarily turned to analytical or data-driven methods~\cite{bohg2013data}.

\begin{figure}[t!]
    \centering
    \includegraphics[width=\linewidth]{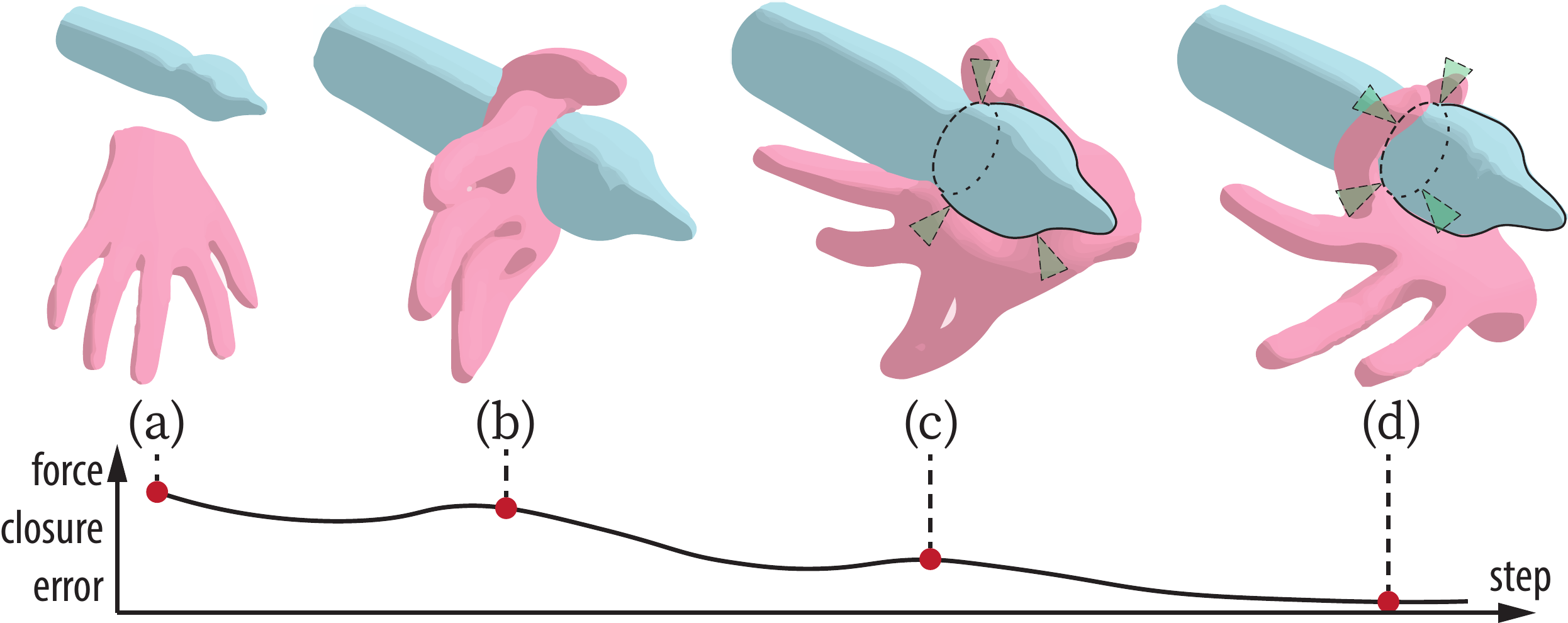}
    \caption{\textbf{Grasp synthesis process by minimizing the force closure error.} The green trianglets in (c)(d) denote the friction cones at contact points used to calculate force closure.}
    \label{fig:intro}
\end{figure}

Analytical methods use manually derived algorithms. Due to the intrinsic complexity of the grasp synthesis, these methods~\cite{ponce1993characterizing,ponce1997computing,li2003computing} typically perform only in limited settings (usually on power grasps~\cite{feix2015grasp}) and are only applicable to specific robotic hand structures. Modern approaches focus more on data-driven methods~\cite{taheri2020grab,karunratanakul2020grasping}, which rely on large datasets of human demonstrations. Although these methods can reproduce (and even interpolate) similar but different grasps compared to human demonstrations, they are inherently difficult to generalize (especially to extrapolate) to arbitrary hand kinematics and unseen grasp types. Furthermore, these data-driven methods usually do not consider the physical stability in producing grasps, making them difficult to deploy on physical robots.

In this paper, we rethink the grasp synthesis problem and derive a differentiable force closure estimator, computed within \textbf{milliseconds} on modern desktops. Such fast computation of force closure opens a new venue for grasp synthesis. Since it does not rely on training data or restrict to specific robotic hand structures, our method can be applied to arbitrary hand structures to synthesize physically stable and diverse grasps.

Specifically, our method is on the basis of two simple yet reasonable and effective assumptions: \textbf{zero friction} and \textbf{equal magnitude of contact forces}, which avoid solving the contact forces as an optimization problem. Intuitively, such assumptions indicate that the contact force on each contact point becomes simply the object's surface normal on that point. Consequently, the overall nested optimization problem is converted to minimizing the errors that violate the above assumptions; see an example in \cref{fig:intro}. In a series of experiments, we demonstrate that our estimated error reflects the difference between surface normal vectors and force closure contact force vectors. We further devise a grasp energy function based on the estimated force closure and validate the force-closure grasp synthesis by minimizing the energy function in six settings.

This paper makes three primary contributions:
(i) We formulate a fast and differentiable estimator of force closure, computed within milliseconds.
(ii) We synthesize diverse grasps with arbitrary hand structures without any training data.
(iii) Since our method is independent of specific hand structures, grasping and manipulation algorithms built upon our method would be easily transferable among competitions and benchmarks that require different end-effectors.

\section{Related Work}

\textbf{Grasp synthesis} literature can be roughly categorized into two schools of thought: analytic and data-driven approach.

The analytic approach generates grasps by considering kinematics and physics constraints~\cite{sahbani2012overview}. Although force closure has been commonly adopted as the physics constraint~\cite{rodriguez2012caging,prattichizzo2012manipulability,rosales2012synthesis,murray2017mathematical}, primary efforts focus on simplifying the search space (\eg, \cite{ponce1993characterizing,ponce1997computing,li2003computing}) as testing force closure is expensive. However, these methods are only effective in specific settings or applications.

The data-driven approach leverages recent advancements in machine learning to estimate grasp points. Despite promising progress~\cite{saxena2008robotic,mahler2017dex,levine2018learning}, this approach relies heavily on large datasets to learn successful grasps, with a particular focus on grippers with limited DoF. Although recent literature~\cite{lin2015robot,brahmbhatt2020contactgrasp,grady2021contactopt} extends this approach to more complex hand models, it still relies on the expensive and tedious collection of human demonstration data. Fundamentally, it is non-trivial for a data-driven approach to generalize the learned model to other hand kinematics. 

An example that does not fall into either of the above categories is the popular toolkit of GraspIt!~\cite{miller2004graspit}. It generates grasps by initializing hand pose randomly, squeezing the fingers as much as possible, and ranking them by a user-defined grasp metric (\eg, a force closure metric). Although this method can generate valid grasps, it is highly inefficient and incapable of generating diverse grasps~\cite{corona2020ganhand}.

A \textbf{force-closure} grasp is a grasp with contact points $\{x_i\in \mathbb{R}^3, i=1,...,n\}$ such that $\{x_i\}$ can resist arbitrary external wrenches with contact forces $f_i$, where $f_i$ lies within the friction cones rooted from $x_i$. The angles of the friction cones are determined by the surface friction coefficient: The stronger the friction, the wider the cone. The force-closure metric is, therefore, irrelevant to the actual hand pose, but only relevant to the contact points and friction cones.
To test whether a set of contact points form a force-closure grasp, the first step is solving an optimization problem regarding contact forces rooted from the points~\cite{boyd2007fast,han2000grasp}. Although various methods have been devised, they all require iterations to jointly solve an auxiliary function, \eg, a support function~\cite{zheng2009distance}, a bilinear matrix inequality~\cite{dai2018synthesis}, or a ray shooting problem~\cite{liu1999qualitative}. As such, solving force-closure grasps under the constraint of hand kinematics and force closure is a nested optimization problem. 

Various methods were proposed to fast approximate this optimization problem, including friction cone approximation with ellipsoids~\cite{tsuji2009easy} and data-driven force closure estimation~\cite{mahler2017dex}. The former method is conceptually similar to this paper but would slow down exponentially as the number of contact points increases, whereas our method is more computationally robust (see \cref{sec:runtime}). The latter method depends heavily on training data and thus suffers from the same problems as data-driven grasp synthesis algorithms do. 

Human grasps, organized into a \textbf{grasp taxonomy}~\cite{feix2015grasp}, provide different levels of power and precision. Most existing grasp synthesis methods focus on synthesizing power grasp, either analytical~\cite{rodriguez2012caging,prattichizzo2012manipulability,rosales2012synthesis,murray2017mathematical} or data-driven~\cite{karunratanakul2020grasping}. At a high cost of annotating object-centric grasp contact information, some data-driven approaches~\cite{brahmbhatt2020contactgrasp,taheri2020grab} demonstrate a certain level of capability to generate a broader range of grasp types.

\section{Differentiable Force Closure}

Formally, given a set of $n$ contact points $\{x_i\in \mathbb{R}^3, i=1,...,n\}$ and their corresponding friction cones $\{(c_i, \mu)\}$, where $c_i$ is the friction cone axis and $\mu$ is the friction coefficient, a grasp is in \emph{force closure} if there exists contact forces $\{f_i\}$ at $\{x_i\}$ within $\{(c_i,\mu)\}$ such that $\{x_i\}$ can resist arbitrary external wrenches. We follow the notations in Dai \etal~\cite{dai2018synthesis} to define a set of contact forces to be force closure if it satisfies the following constraints:
\begin{subequations}
    \begin{align}
        GG' &\succeq \epsilon I_{6\times 6}, \label{eq_c1}\\
        Gf &= 0, \label{eq_c2}\\
        f_i^T c_i &> \frac{1}{\sqrt{\mu^2+1}}|f_i|, \label{eq_c3}\\
        x_i &\in S, \label{eq_c4}
    \end{align}
    \label{eq_c}%
\end{subequations}%
where $S$ is the object surface, and
\begin{align}
    G &= \begin{bmatrix}
    I_{3\times 3} & I_{3\times 3} & ... & I_{3\times 3}\\
    \lfloor x_1 \rfloor_\times & \lfloor x_2 \rfloor_\times & ... & \lfloor x_n \rfloor_\times
    \end{bmatrix}, \label{eq_G}\\
    \lfloor x_i \rfloor_\times &= \begin{bmatrix}
    0 & -x_i^{(3)} & x_i^{(2)}\\
    x_i^{(3)} & 0 & -x_i^{(1)}\\
    -x_i^{(2)} & x_i^{(1)} & 0
    \end{bmatrix}.
\end{align}

The form of $\lfloor x_i \rfloor_\times$ ensures the cross product $\lfloor x_i \rfloor_\times f_i = x_i \times f_i$, where $f=[f_1^T f_2^T ... f_n^T]^T\in\mathbb{R}^{3n}$ is the unknown variable of contact forces. In \cref{eq_c1}, $\epsilon$ is a small constant. $A \succeq B$ means $A-B$ is positive semi-definite, \ie, it is symmetric, and all its eigenvalues are non-negative. \cref{eq_c1} states that $G$ is full rank. \cref{eq_c2} states that the contact forces cancel out each other so that the net wrench is zero. \cref{eq_c3} prevents $f_i$ from deviating from the friction cone $\{(c_i,\mu)\}$. \cref{eq_c4} constrains contact points to be on the object surface. 

\textbf{Relaxation:}
Of note, \cref{eq_c2} is bilinear on $x_i$ and $f_i$. Given a set of contact points $\{x_i\}$, verification of force closure requires finding a solution of $\{f_i\}$. The time complexity for computing such a solution is linear \wrt the number of contact points~\cite{dai2018synthesis}.
Here, we rewrite \cref{eq_c2} to
\begin{subequations}
    \begin{align}
        Gf = G(f_n+f_t) = 0,\\
        G\frac{f_n}{\Vert f_n\Vert _2} = -\frac{Gf_t}{\Vert f_n\Vert _2},\\
        Gc = -\frac{Gf_t}{\Vert f_n\Vert _2},
    \end{align}
    \label{eq_3c}%
\end{subequations}%
where $f_n$ and $f_t$ are the normal and tangential components of contact force $f$ in the force closure model, and $c=[c_1^T c_2^T ... c_n^T]^T$ is the set of friction cone axes. We obtain $c_i$ as the surface normal of the object on $x_i$, which is easily accessible in many shape representations. We use $Gc$ to approximate $Gf$, and therefore relax \cref{eq_c} to 
\begin{subequations}
    \begin{align}
        GG' &\succeq \epsilon I_{6\times 6}, \label{eq_2c1}\\
        \Vert Gc\Vert_2 & < \delta, \label{eq_2c2}\\
        x_i &\in S, \label{eq_2c3}
    \end{align}
    \label{eq_2c}%
\end{subequations}%
where $\delta$ is the maximum allowed error introduced from our relaxation. By adopting \cref{eq_2c}, we no longer need to solve the unknown variable $f$. The constraints of $x_i$ becomes quadratic. Hence, the verification of force closure can now be computed extremely fast. The residual in $\Vert Gc\Vert_2$ reflects the difference between contact forces and friction cone axes. 

To allow gradient-based optimization, we further cast \cref{eq_2c} as a soft constraint in the form
\begin{equation}
    FC(x,O) = \lambda_0^-(GG' - \epsilon I_{6\times6}) + \Vert Gc\Vert _2 + w\sum_{x_i\in x}d(x_i,O),
    \label{eq_fc_soft}
\end{equation}
where $\lambda_0^-(\cdot)=\mathrm{ReLU}(-\lambda_0(\cdot))$ gives the negative part of the smallest eigenvalue of a matrix, and $d(x,O)$ returns the distance from point $x$ to the surface of object $O$. The scalar $w$ controls the weight of the distance term. By minimizing the three terms, we are looking for $\{x_i\}$ that satisfies the three constraints in \cref{eq_2c}, respectively.

\textbf{Implications of Assumptions:}
Using surface normal vectors to approximate contact forces implies zero friction and equal magnitude contact forces. Such an assumption may \textit{seem} to eliminate a large pool of force-closure contact-point compositions. In practice, however, this is not the case: A residual in $\Vert Gc\Vert _2$ indicates that the existence of friction $f_t$ and difference in force magnitude $f_n$ on contact forces. By allowing the residual to be smaller than a reasonable threshold $\delta$, we allow the tangential and normal components of the contact forces to deviate within a reasonable range.

\begin{figure}[t!]
    \centering
    \includegraphics[width=\linewidth]{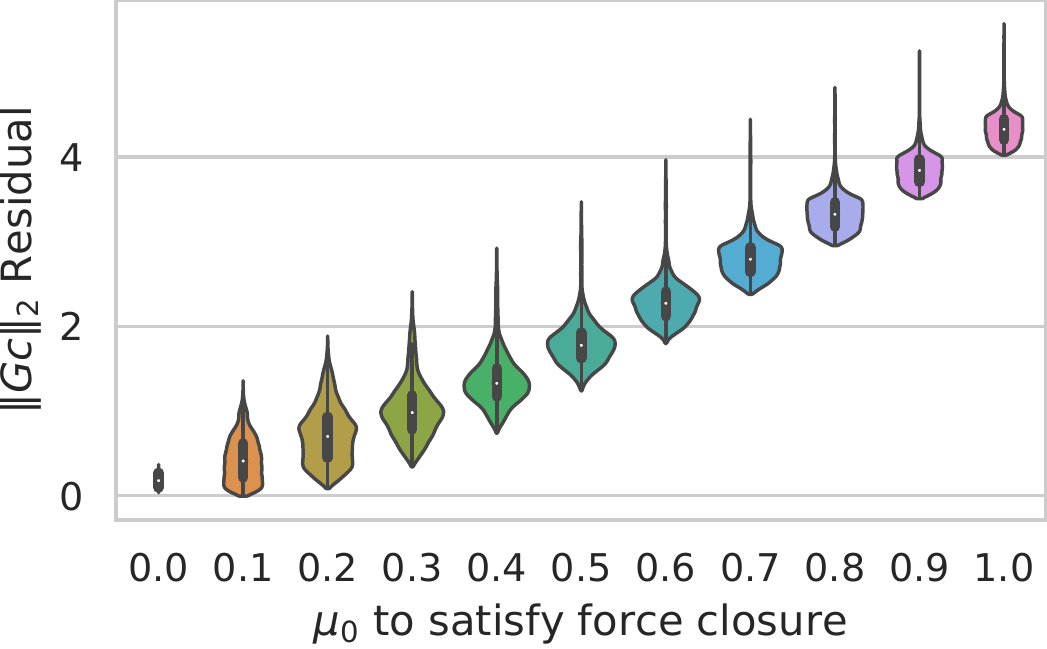}
    \caption{\textbf{$\mathbf{\Vert Gc\Vert_2}$ residual from \cref{eq_2c2} (y-axis) against minimum friction coefficient $\mathbf{\mu_0}$ (x-axis).} The violinplot is the distribution of $\Vert Gc\Vert_2$ residuals of all examples that require a minimum friction coefficient $\mu_0$ to pass the classic force closure test. Overall, these two values are linearly correlated; $\Vert Gc\Vert_2\approx 4.035\mu_0$.}
    \label{fig:estimate_vs_mu}
\end{figure}

To further verify our interpretation, we randomly sample 500,000 grasps, each containing three contact points on the surface of a unit sphere. For each grasp, we compute the \textit{minimum} friction coefficient $\mu_0$ required for the grasp to satisfy the classic force closure constraints described in \cref{eq_c}. \cref{fig:estimate_vs_mu} plots the residual $\Vert Gc\Vert_2$ from \cref{eq_2c2} against $\mu_0$; it shows an almost linear relation between $\mu_0$ and $\Vert Gc\Vert_2$.

\begin{figure}[t!]
    \centering
    \includegraphics[width=\linewidth]{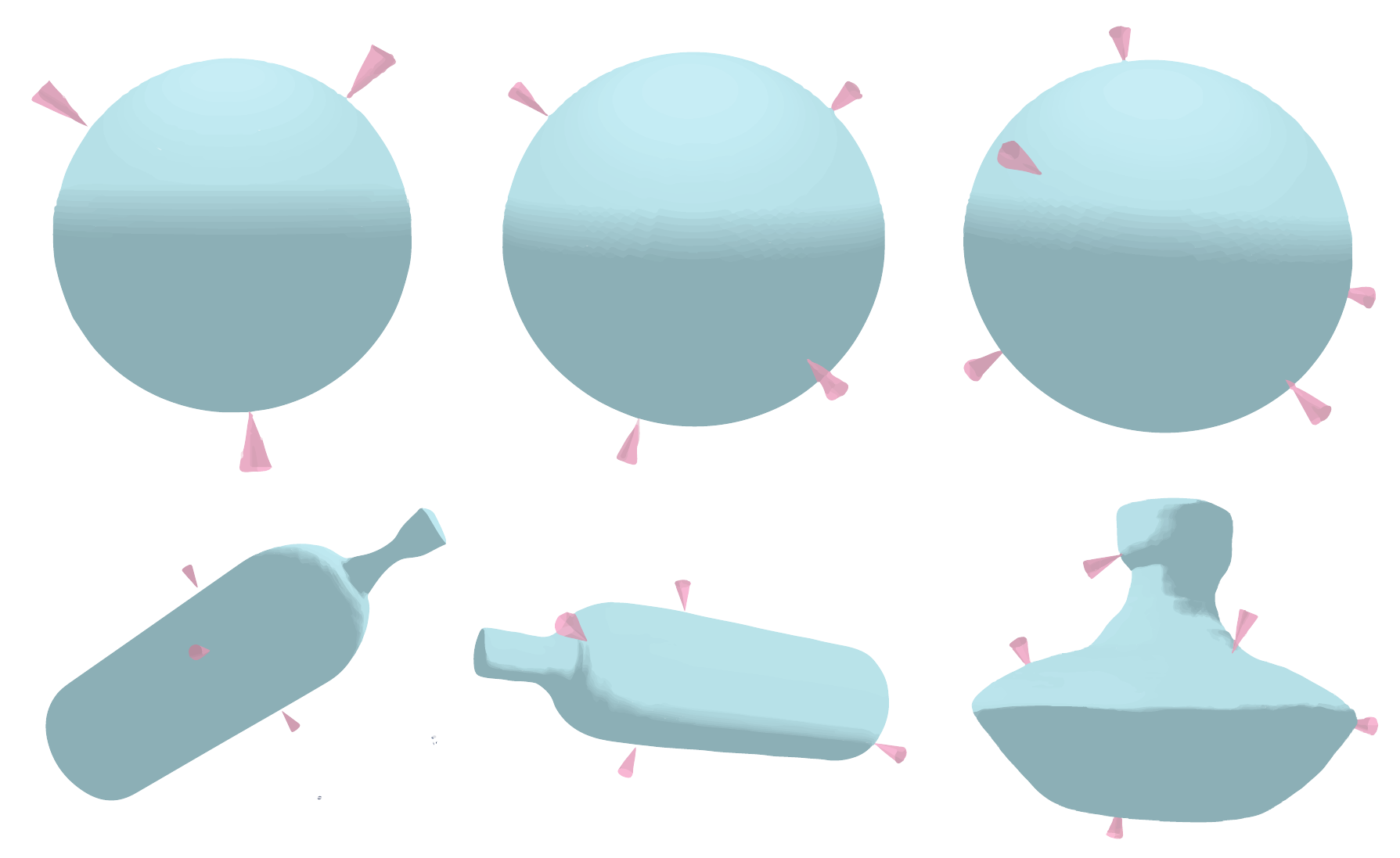}
    \caption{\textbf{Force-closure contact-point generations on unit spheres (top) and daily objects (bottom) by minimizing \cref{eq_fc_soft}.} Objects in each columns have 3, 4, and 5 contact points, respectively.}
    \label{fig:force_closure}
\end{figure}

\textbf{Force-closure Contact-point Generation:}
By directly minimizing the soft force closure constraint, we can synthesize force closure contact points with \textit{arbitrary} shapes. Specifically, we run gradient descent on contact point positions $x$ to minimize \cref{eq_fc_soft}. \cref{fig:force_closure} shows the computed contact points on a unit sphere and some daily objects. Despite our assumptions, minimizing our force closure estimation can indeed properly generate force-closure contact points.

\section{Grasp Synthesis}

In this section, we first describe how to leverage our differentiable force closure estimator to formulate a probability distribution of grasping. Next, we devise an optimization algorithm to sample diverse grasps from the distribution.

\textbf{Formulation:}
We formulate the grasp synthesis problem as sampling from a conditional Gibbs distribution:
\begin{equation}
    P(H|O) = \frac{P(H,O)}{P(O)} \propto P(H,O) = \frac{1}{Z}\exp^{- E(H,O)},
    \label{eq:p}
\end{equation}
where $Z$ denotes the intractable normalizing constant, $H$ the hand, $O$ the object, and $E(H,O)$ the energy function. We rewrite $E(H,O)$ as the minimum value of the energy function $E_\mathrm{grasp}(H,x,O)$ \wrt contact point choices $x$:
\begin{equation}
    \begin{aligned}
        & E(H,O) = \min_{x\subset S(H)} E_\mathrm{grasp}(H,x,O)\\
        = & \min_{x\subset S(H)}FC(x,O) + E_\mathrm{prior}(H) + E_\mathrm{pen}(H,O),
        \label{eq_energy}
    \end{aligned}
\end{equation}
where $S(H)$ is a set of points sampled uniformly from the surface of a hand with pose $H$. We denote the selected contact points from hand surface as $x\subset S(H)$. $FC(x,O)$ is the soft constraint from \cref{eq_fc_soft}. $E_\mathrm{prior}(H)$ is the \textit{energy prior} of the hand pose. Its exact form depends on the hand definition. The \textit{penetration energy} is defined as $E_\mathrm{pen}(H,O)=\sum_{v\in S(H)}\sigma(v,O)$, where $\sigma(v,O)$ is a modified distance function between a point $v$ and an object $O$:
\begin{equation}
    \begin{aligned}
        \sigma(v,O)= 
        \begin{cases} 
        0 &\mbox{ if $v$ outside $O$}\\
        |d| &\mbox{ otherwise}
        \end{cases},
    \end{aligned}
\end{equation}
where $d$ is the distance from $v$ to surface of $O$. 

\textbf{Algorithm:}
Due to the complexity of human hand kinematics, our grasp energy suffers from a complex energy landscape. A na\"ive gradient-based optimization algorithm is likely to stop at sub-optimal local minima. We use a modified \ac{mala} to overcome this issue; see the algorithm details in \cref{algo:mala}. The random walk aspect of Langevin dynamics provides the chance of escaping bad local minima. Our algorithm starts with random initialization of hand pose $H$ and contact points $x\subset S(H)$. Next, we run our algorithm $L$ iterations to update $H, x$ and maximize $P(H,O)$. In each iteration, our algorithm randomly decides to update either the hand pose by Langevin dynamics or one of the contact points to a point uniformly sampled from the hand surface. The updates are accepted or rejected according to the Metropolis-Hastings algorithm, in which a lower-energy update is more likely to be accepted than a higher-energy one. 

\begin{algorithm}[ht!]
    \small
    \caption{Modified \ac{mala} Algorithm}
    \label{algo:mala}
    \SetAlgoLined
    \KwIn{Energy function $E_\mathrm{grasp}$, object shape $O$, step size $\eta$, Langevin steps $L$, switch probability $\rho$}
    \KwOut{grasp parameters $H,x$}
    Initialize $H,x$\\
    \For{$\mathrm{step}=1:L$}{
        \eIf{$\mathrm{rand()} < \rho$}{
            Propose $H^*$ according to Langevin dynamics
            \begin{align*}
                H^*= H - \frac{\eta^2}{2}\frac{\partial}{\partial H}E_\mathrm{grasp}(H,x,O) + \eta\epsilon,
            \end{align*}
            where $\epsilon\sim N(0,1)$ is a Gaussian noise
        }{
            Propose $x^*$ by sampling from $S(H)$\label{step:sample}\\
        }
        Accept $H\gets H^*,x\gets x^*$ by Metropolis-Hastings algorithm using energy function $E_\mathrm{grasp}$
    }
\end{algorithm}

Of note, different compositions of contact points correspond to different grasp types as they contribute to some of the classification basis of the grasp taxonomy, including virtual finger assignment and opposition type. Hence, sampling contact points on \cref{step:sample} in \cref{algo:mala} is crucial for exploring different types of grasps. In practice, we also empirically find that this step is essential for escaping bad local minima.

\section{Simulation}

We detail experimental setup with analysis in simulation. 

\subsection{Simulation Setup}

\paragraph*{Hand Model}

We use MANO~\cite{romero2017embodied} to model the humanoid hand. It is a parameterized 3D hand shape model that maps low-dimensional hand poses to 3D human hand shapes. We use the norm of the PCA weights of the hand pose as $E_\mathrm{prior}(H)$. Since MANO vertices are distributed uniformly across the hand surface, we sample points from the hand surface by directly sampling from MANO vertices.

\paragraph*{Object Model}

We use the DeepSDF model~\cite{park2019deepsdf} to model the objects to be grasped. DeepSDF is a densely connected neural network that implicitly represents the shape surface and estimates the signed distance from a position to an object surface; the signed distance is negative if the point is inside the object, and vice versa. The 0-level set composes the surface of the object. We obtain the object surface normal by taking the derivative of the signed distance \wrt the input position.

We test our grasp synthesis algorithm on various bottles retrieved from ShapeNet dataset~\cite{chang2015shapenet}. Given the pre-trained DeepSDF model of an object, we randomly initialize a MANO hand and use \cref{algo:mala} to sample the hand pose and contact points from $P(H|O)$. We set the step size $\eta=0.1$, switch probability $\rho=0.85$, distance weight $w=1$, and Langevin steps $L=10^6$. We filter out samples in bad local minima by keeping samples that satisfy following empirical constraints:
\begin{subequations}
    \begin{align}
        \Vert Gc\Vert_2 &< 0.5\\
        \sum_{x_i\in x}d(x_i,O)^2 &< 0.02\\
        E_\mathrm{pen}(H,O) &< 0.02
    \end{align}%
    \label{eq:empirical}%
\end{subequations}%
where $x$ is the set of contact points on the hand surface, and $c$ the friction cone axes at contact points.

\subsection{Runtime Analysis}\label{sec:runtime}

\begin{figure}[t!]
    \centering
    \includegraphics[width=\linewidth]{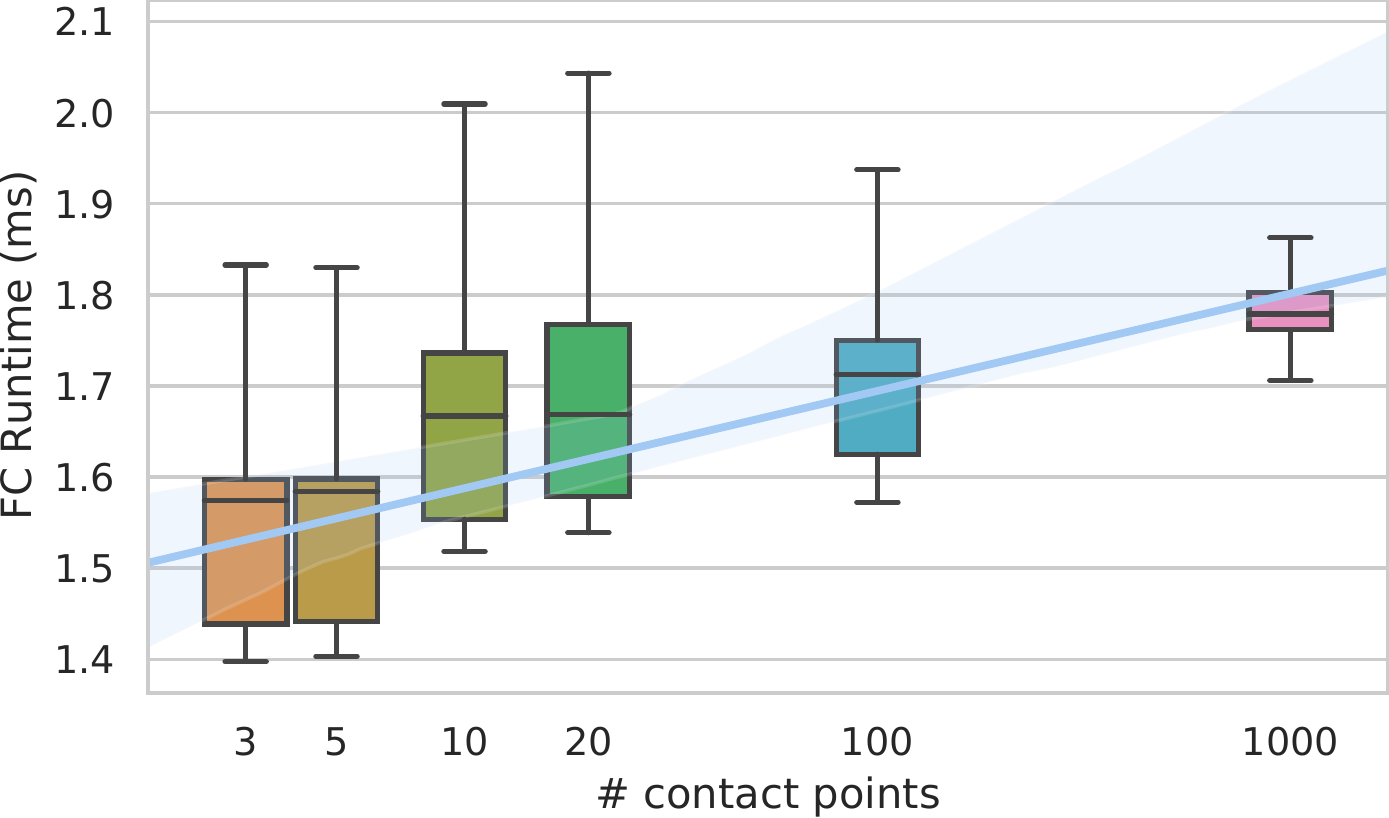}
    \caption{\textbf{Boxplot and log-linearly fitted curve of the runtime of \cref{eq_fc_soft} \wrt to the number of contact points.} We run a simulated test of force closure with 3, 5, 10, 20, 100, and 1000 contact points for 1,000 iterations. X-axis is the number of contact points in log scale. Y-axis is the runtime of our force closure error estimate. The shaded area denotes the 95\% confidence interval. The light blue line denotes that the estimated relation between the FC runtime $t$ and the number of contact points $n$ is $t\approx 0.107\times\log_{10}n + 1.502$.}
    \label{fig:fc_time}
\end{figure}

\cref{fig:fc_time} shows the time complexity of testing force closure using \cref{eq_fc_soft} and our simulation setup, wherein we further fit a log-linear curve of the running time \wrt the number of contact points. Each test takes 1-2ms to run on an NVIDIA 3090 GPU, significantly faster than the exact solution~\cite{dai2018synthesis}. We also observe that roughly 80\% of the total runtime is spent at the computation of surface normal; this operation is particularly slow because it takes a derivative of the DeepSDF model. Taken together, these empirical results in simulation indicate that a further improvement in runtime efficiency would be achievable with a more computationally tractable object shape representation, an interesting future research direction.

\begin{figure}[t!]
    \captionsetup[subfigure]{justification=centering}
    \begin{subfigure}{0.32\linewidth}
        \centering
        \includegraphics[width=\textwidth]{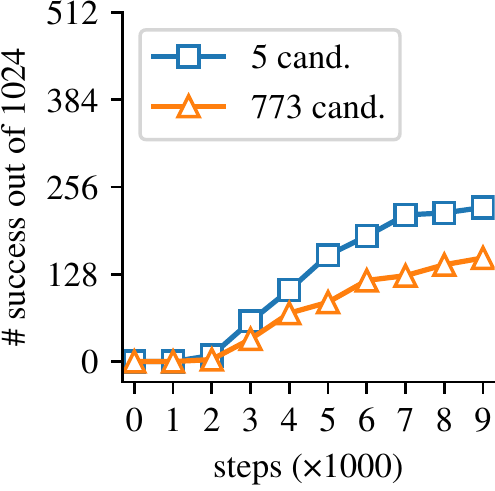}
        \caption{}
    \end{subfigure}%
    \hfill
    \begin{subfigure}{0.32\linewidth}
        \centering
        \includegraphics[width=\textwidth]{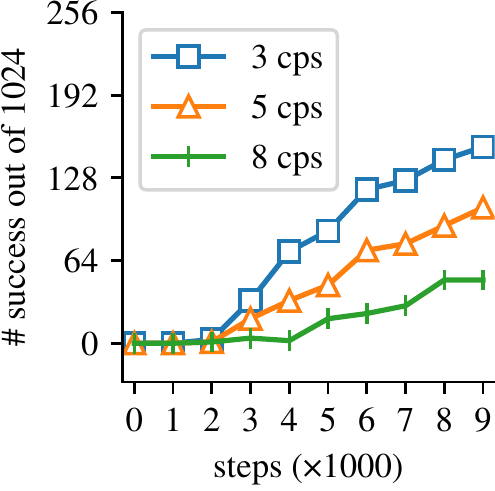}
        \caption{}
    \end{subfigure}%
    \hfill
    \begin{subfigure}{0.32\linewidth}
        \centering
        \includegraphics[width=\textwidth]{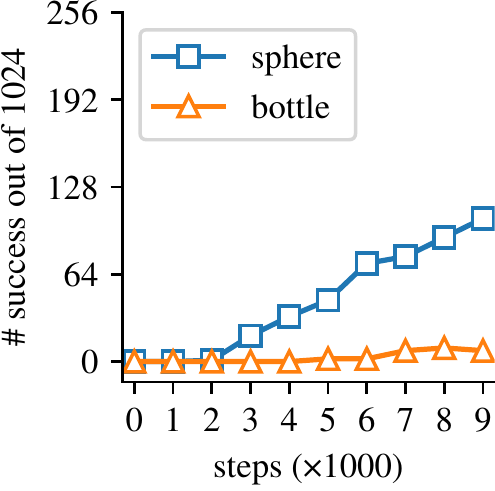}
        \caption{}
    \end{subfigure}%
    \caption{\textbf{Runtime analysis of grasp synthesis.} (a) Grasp spheres with 3 contact point candidates (one on each fingertip) and 773 candidates (uniformly distributed over entire hand surface). (b) Grasp spheres with 3, 5, and 8 contact points. (c) Grasp spheres and ShapeNet bottles with 3 contact points.}
    \label{fig:grasp_time}
\end{figure}

We further examine the efficiency of \cref{algo:mala} under different settings in \cref{fig:grasp_time}. We synthesize 512 examples for $10^4$ steps under each setting and count the number of successful synthesis results that satisfy \cref{eq:empirical} at each step. We observe that the algorithm is more likely to succeed with a smaller search space of contact point selection (less contact points and less candidates) and with simpler object shapes. For complex shapes such as bottles, our method only produced 5 successful syntheses after $10^4$ steps, and the first acceptable synthesis emerged at 1047th step. On average, each step takes 224.4ms for bottles, of which over 200ms is spent on computing the gradient of the force closure estimate---it involves computing the second-order derivative of the DeepSDF function.

\subsection{Refinement}

While our modified \ac{mala} algorithm can produce realistic results, we still observe physical inconsistencies in the synthesized examples such as penetrations and gaps between contact points and object surface. To resolve these issues, we further refine the synthesized results by minimizing $E_\mathrm{grasp}$ using gradient descent on $H$. We do not update the contact point selection $x$ in this step, since we hope to focus on optimizing the physical consistency in this step rather than exploring the grasp landscape for diverse grasp types.

\section{Results}

We showcase our method's capabilities in simulation.

\subsection{Grasp Synthesis}

\begin{figure*}[t!]
    \captionsetup[subfigure]{justification=centering}
    \hfill
    \begin{subfigure}{0.15\linewidth}
        \centering
        \begin{overpic}[width=\linewidth,height=2.5cm,keepaspectratio]{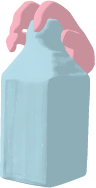}
            \put(-70,42){\color{black}\linethickness{0.5mm}%
                \fontsize{10}{10}\selectfont{}without
            }%
            \put(-70,30){\color{black}\linethickness{0.5mm}%
                \fontsize{10}{10}\selectfont{}refinement
            }%
        \end{overpic}
        \caption{\\\textbf{FC=0}\\SD=0.0143}
    \end{subfigure}%
    \begin{subfigure}{0.15\linewidth}
        \centering
        \includegraphics[width=\linewidth,height=2.5cm,keepaspectratio]{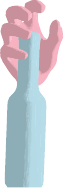}
        \caption{\\\textbf{FC=0}\\SD=0.0457}
    \end{subfigure}%
    \begin{subfigure}{0.15\linewidth}
        \centering
        \includegraphics[width=\linewidth,height=2.5cm,keepaspectratio]{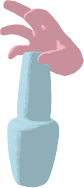}
        \caption{\\\textbf{FC=0.0467}\\SD=0.0323}
    \end{subfigure}%
    \begin{subfigure}{0.15\linewidth}
        \centering
        \includegraphics[width=\linewidth,height=2.5cm,keepaspectratio]{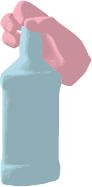}
        \caption{\\FC=0.0581\\SD=0.0128}
    \end{subfigure}%
    \begin{subfigure}{0.15\linewidth}
        \centering
        \includegraphics[width=\linewidth,height=2.5cm,keepaspectratio]{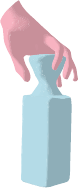}
        \caption{\\FC=0.1035\\SD=0.0274}
    \end{subfigure}%
    \begin{subfigure}{0.15\linewidth}
        \centering
        \includegraphics[width=\linewidth,height=2.5cm,keepaspectratio]{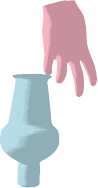}
        \caption{\\FC=1.2294\\SD=0.0053}
    \end{subfigure}%
    
    \hfill
    \begin{subfigure}{0.15\linewidth}
        \centering
        \begin{overpic}[width=\linewidth,height=2.5cm,keepaspectratio]{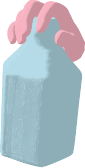}
            \put(-70,42){\color{black}\linethickness{0.5mm}%
                \fontsize{10}{10}\selectfont{}with
            }%
            \put(-70,30){\color{black}\linethickness{0.5mm}%
                \fontsize{10}{10}\selectfont{}refinement
            }%
        \end{overpic}
        \caption{\\\textbf{FC=0}\\\textbf{SD=0.0033}}
    \end{subfigure}%
    \begin{subfigure}{0.15\linewidth}
        \centering
        \includegraphics[width=\linewidth,height=2.5cm,keepaspectratio]{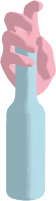}
        \caption{\\\textbf{FC=0}\\\textbf{SD=0.0022}}
    \end{subfigure}%
    \begin{subfigure}{0.15\linewidth}
        \centering
        \includegraphics[width=\linewidth,height=2.5cm,keepaspectratio]{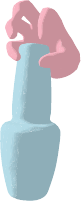}
        \caption{\\FC=0.0900\\\textbf{SD=0.0015}}
    \end{subfigure}%
    \begin{subfigure}{0.15\linewidth}
        \centering
        \includegraphics[width=\linewidth,height=2.5cm,keepaspectratio]{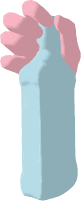}
        \caption{\\\textbf{FC=0}\\\textbf{SD=0.0020}}
    \end{subfigure}%
    \begin{subfigure}{0.15\linewidth}
        \centering
        \includegraphics[width=\linewidth,height=2.5cm,keepaspectratio]{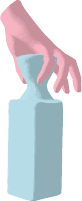}
        \caption{\\\textbf{FC=0}\\\textbf{SD=0.0006}}
    \end{subfigure}%
    \begin{subfigure}{0.15\linewidth}
        \centering
        \includegraphics[width=\linewidth,height=2.5cm,keepaspectratio]{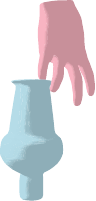}
        \caption{\\\textbf{FC=1.1509}\\\textbf{SD=0.0004}}
    \end{subfigure}%
    \caption{\textbf{Examples of synthesized grasps.} Top: synthesized grasps before refinement. Bottom: the same set of synthesized grasps after refinement. FC: estimated force closure error. SD: mean distance from each contact point to the object surface. Left to right: examples with zero FC error, small FC error, and high FC error qualitatively illustrate how our estimation of force closure correlates to grasp quality.}
    \label{fig:ablation}
\end{figure*}

\cref{fig:ablation} shows synthesis results with and without the refinement step: Higher values of our force closure estimation corresponds to non-grasps, whereas force closure estimations close to zero are as good as the ones with force closure estimations equal to zero. \textbf{This observation confirms our previous analysis.} We also notice cases when the synthesis is trapped in bad local minima; we show two examples in the last column of \cref{fig:ablation}. These examples exhibit large values in our force closure estimator, which happened due to the non-convexity of the optimization problem; one cannot avoid every bad minimum with gradient-based methods. Fortunately, we can identify these examples by their high force closure scores.

\subsection{Physical Stability}

We verify the physical stability of our synthesized examples by simulating the samples in PyBullet. Specifically, we set gravity to be $[0,0,-10]m/s^{-2}$ and use the default values of friction coefficients in PyBullet. We assumed both the hand and the object to be rigid bodies. An example is deemed to be a successful grasp if the object's vertical drop is less than $0.3m$ after 1000 steps of simulation, or $16.67s$. 

A grasp's physical stability depends on the force closure score of the contact points and whether the contact points are close enough to the object surface. We set two thresholds on the contact point distance; \cref{tbl:success_rate} tabulates detailed comparisons of the success rate between our method against state-of-the-art algorithms: To our best knowledge, Przybylski \etal~\cite{przybylski2010unions} is the state-of-the-art analytic approach, whereas Ottenhaus \etal~\cite{ottenhaus2019visuo} is the state-of-the-art data-driven approach.

\begin{table}[ht!]
    \centering
    \caption{Grasp success rates:\\ours \textit{vs.} state-of-the-art methods}
    \begin{tabular}{cc}
        \toprule
        \textbf{method} & \textbf{success rate}  \\
        \midrule
        Unions of Balls~\cite{przybylski2010unions} & 72.53\%\\
        \textbf{Visuo-Haptic~\cite{ottenhaus2019visuo}} & \textbf{85.00\%}\\
        Ours ($\sigma<0.0015$) & 76.98\%\\
        \textbf{Ours ($\sigma<$0.0005)} & \textbf{85.00\%}\\
        \bottomrule
    \end{tabular}
    \label{tbl:success_rate}
\end{table}

Of note, although Ottenhaus \etal~\cite{ottenhaus2019visuo} reported 95\% success rate in the original paper, many of the objects being tested have simple shapes, such as a sphere or a box; the success rate would drop to 85\% when we remove these simple objects. Additionally, neither of the two state-of-the-art methods has demonstrated the ability to synthesize diverse types of grasps. Although some other data-driven methods have demonstrated a certain level of diverse grasp synthesis, they fail to report their physical stability as it is not their primary focus.

\subsection{Diversity of the Grasp Types}

To evaluate the diversity of the grasps generated by the proposed method, we examine the energy landscape of our grasp energy function; we use ADELM algorithm~\cite{hill2019building} to build the energy landscape mapping of our grasps energy function $E(H,O)$. Below, we show that grasps defined by our energy function loosely aligns with the carefully defined grasps taxonomy~\cite{feix2015grasp} when applied to humanoid hands.

\begin{figure*}[t!]
    \centering
    \includegraphics[width=\linewidth]{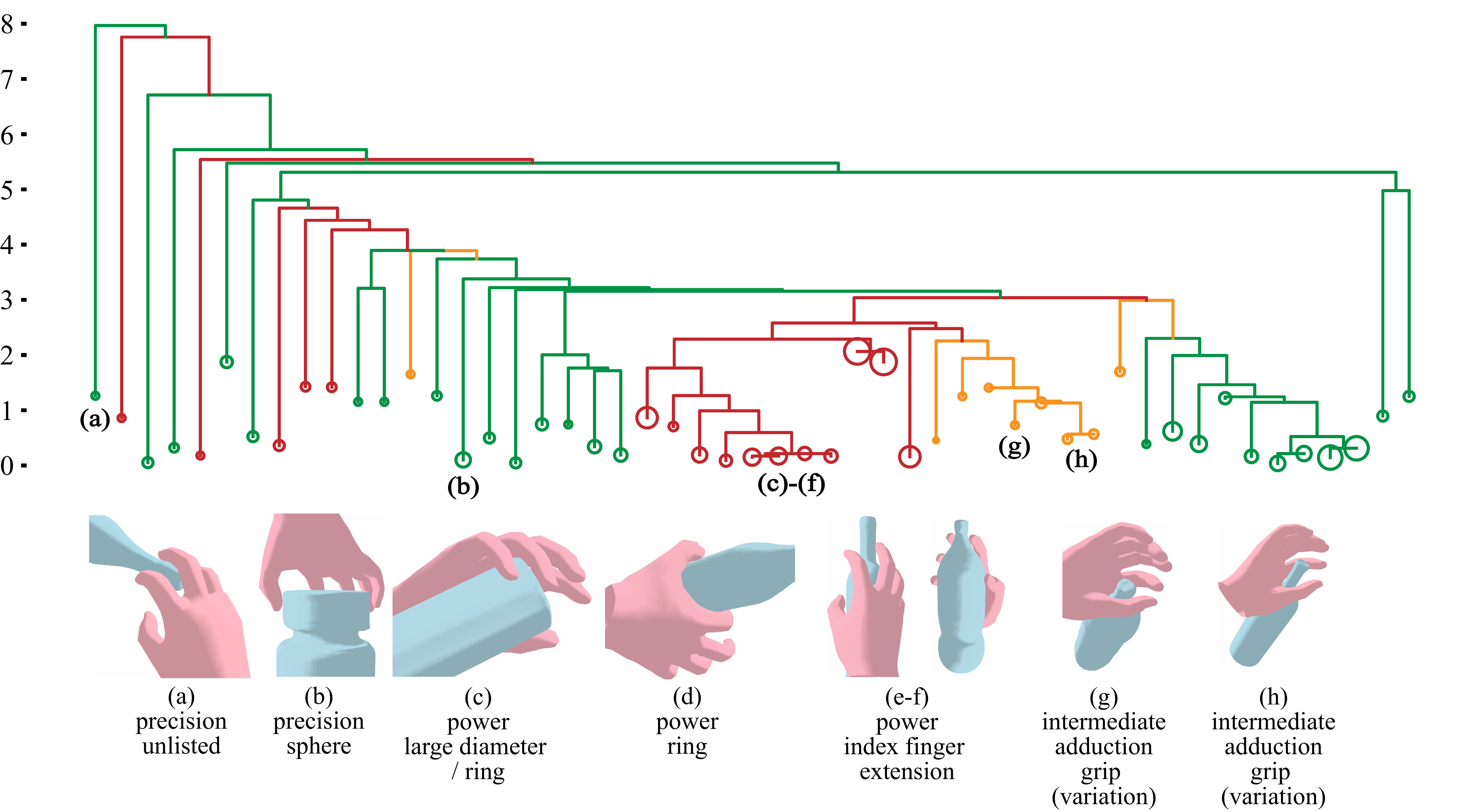}
    \caption{\textbf{Energy landscape mapping generated by the ADELM algorithm~\cite{hill2019building}; best viewed in color.} Top: disconnectivity diagram of the energy landscape of our energy function $E(H,O)$. Green minima denote precision grasps, red power grasps, and yellow intermediate grasps. Bottom: examples from selected local minima; minima with lower energy barriers in between have similar grasps. We also label the grasp taxonomy of each example according to Feix \etal~\cite{feix2015grasp}. Examples marked as \emph{unlisted} do not belong to any manually classified type.}
    \label{fig:landscape}
\end{figure*}

Specifically, we collected 371 synthesized grasp examples and adopted the ADELM algorithm~\cite{hill2019building} to find \ac{mep} between them. We project the \ac{mep} between examples to a disconnectivity graph in \cref{fig:landscape}. In the disconnectivity graph, each circle at the bottom represents a local minima group. The size of the circle indicates how many synthesized examples fall into this group. The height of the horizontal bar between two groups represent the maximum energy (or energy barrier) along the \ac{mep} between two groups. The \ac{mep} with lowest barriers connect smaller groups into larger groups, and this process is repeated until all examples are connected. The produced disconnectivity graph is an estimation of the true landscape of the energy function. Energy landscape mapping in \cref{fig:landscape} shows that the local minima with low energy barriers between them have similar grasps, and those with high energy barriers between them tend to have different grasps. We also observe that the energy landscape contains all three categories in the power and the precision dimension as described in Feix \etal~\cite{feix2015grasp}. 

\begin{figure*}[t!]
    \centering
    \begin{subfigure}[t]{0.32\linewidth}
        \includegraphics[width=\linewidth]{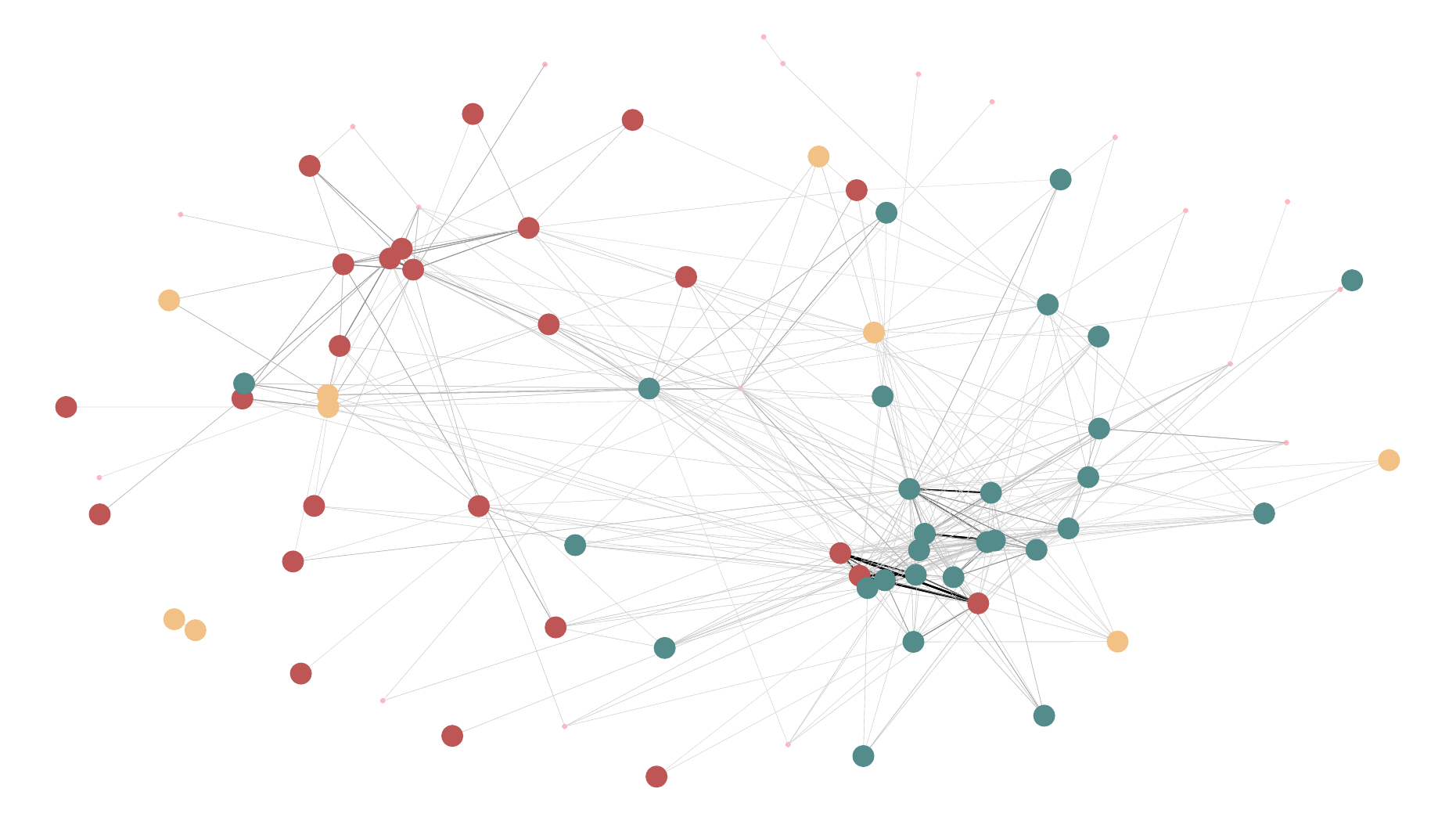}
        \caption{\textcolor{myred}{Red}: power grasps. \textcolor{myyellow}{Yellow}: intermediate grasps. \textcolor{mygreen}{Green}: Precision grasps. Other: Unlisted.}
        \label{fig:tax-a}
    \end{subfigure}%
    \hfill
    \begin{subfigure}[t]{0.32\linewidth}
        \includegraphics[width=\linewidth]{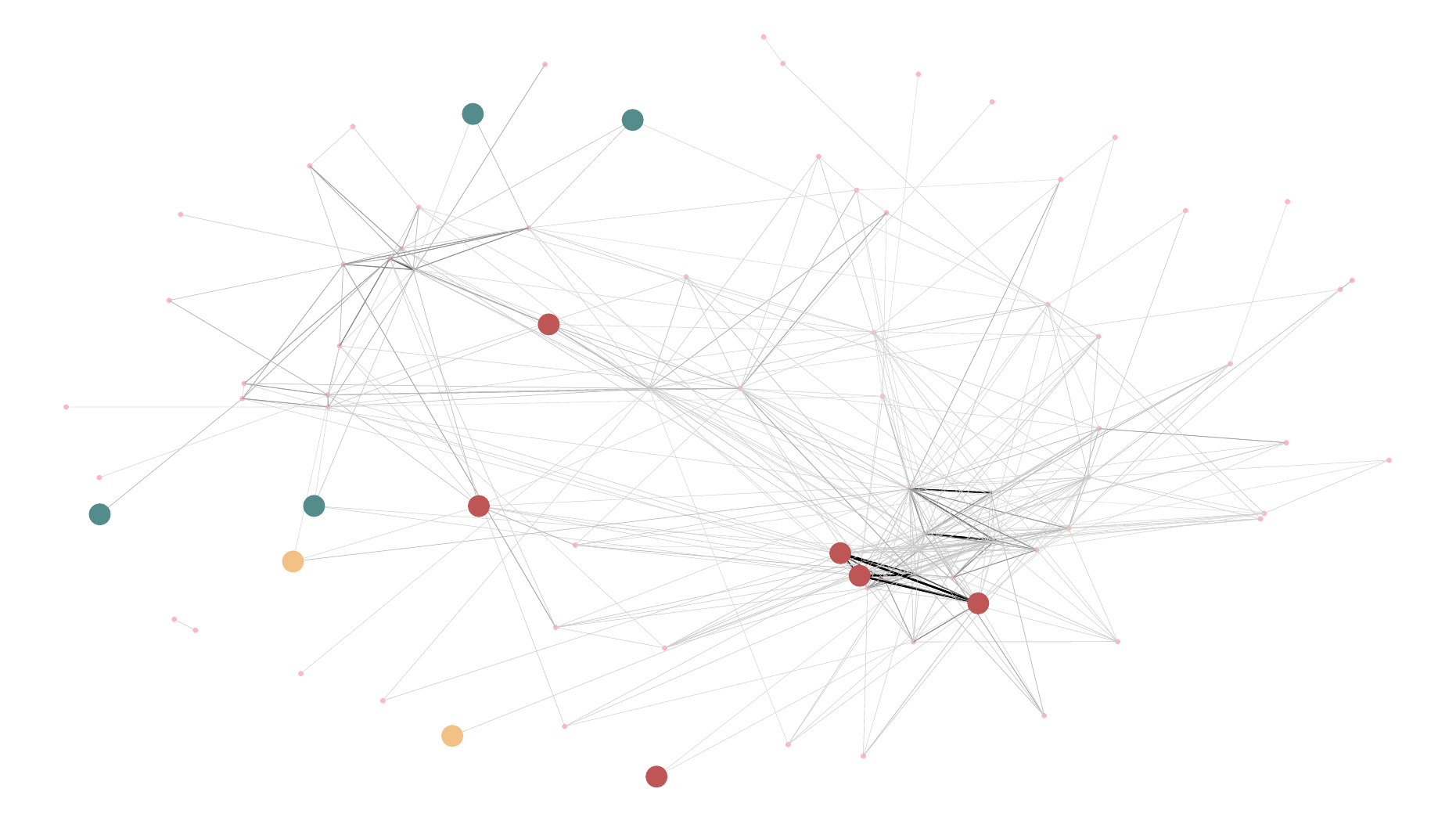}
        \caption{\textcolor{myred}{Red}: power sphere grasps. \textcolor{myyellow}{Yellow}: power disk grasp. \textcolor{mygreen}{Green}: power cylinder grasps (large diameter, medium wrap, small diameter)}
        \label{fig:tax-b}
    \end{subfigure}%
    \hfill
    \begin{subfigure}[t]{0.32\linewidth}
        \includegraphics[width=\linewidth]{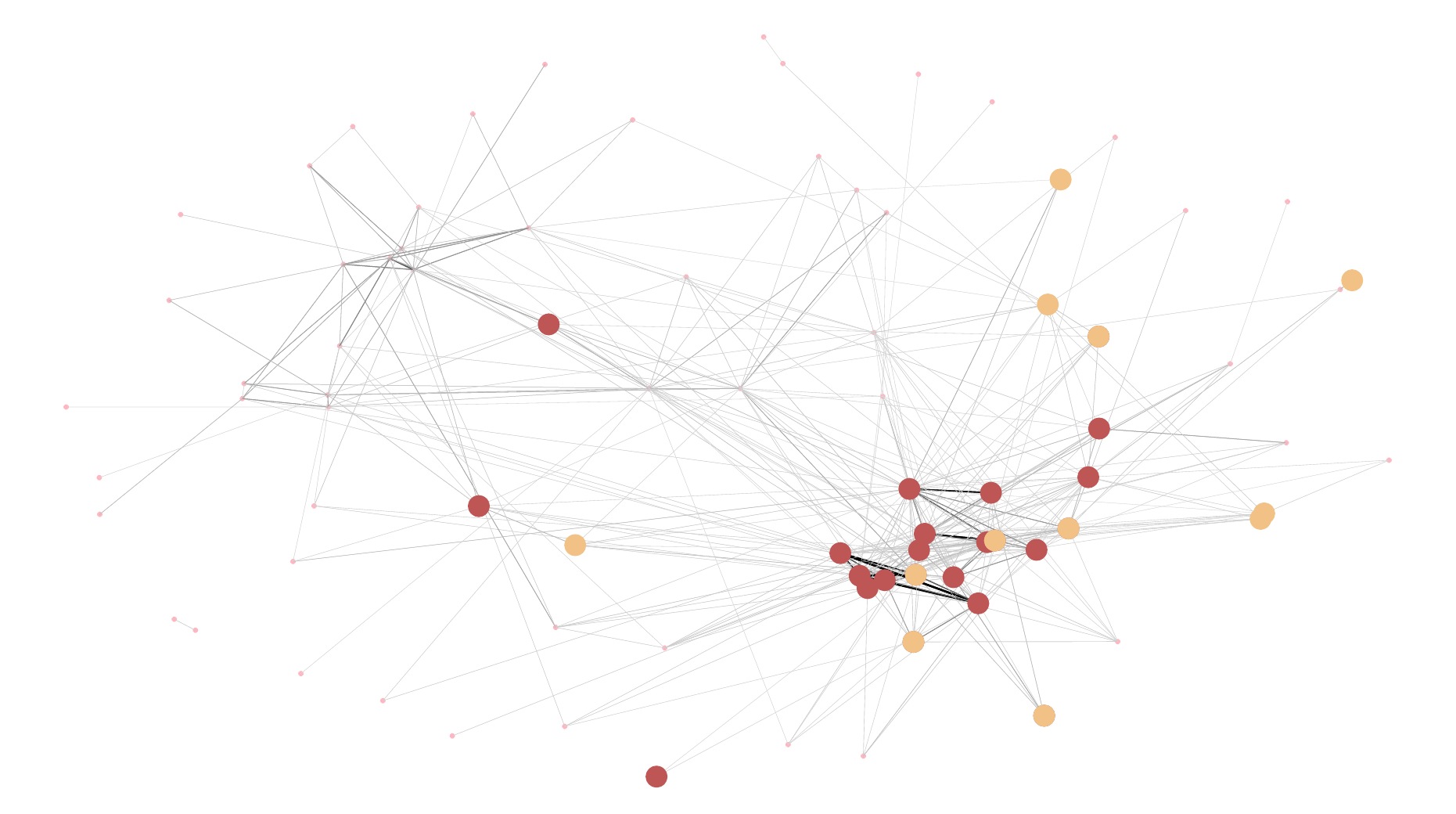}
        \caption{\textcolor{myred}{Red}: power/precision sphere grasps. \textcolor{myyellow}{Yellow}: tri/quad-pod grasps.}
        \label{fig:tax-c}
    \end{subfigure}%
    \caption{Alignment between our energy landscape and existing grasp taxonomy \cite{feix2015grasp}; best viewed in color.}
    \label{fig:taxonomy}
\end{figure*}

To provide a more comprehensive understanding of the alignment between our energy landscape and the existing taxonomy, we further plot the local minima groups as a 2D graph in \cref{fig:taxonomy}, which supplements the 1D energy landscape shown in \cref{fig:landscape}. In \cref{fig:taxonomy}, each node represents a local minima group. We arrange the nodes so that groups with low energy barriers between them are placed closer to each other. The edges between nodes also indicate the energy barriers: Thicker edges indicate lower barriers, and no edge between two nodes means no pathway between the two groups is formed.

Specifically, \cref{fig:tax-a} shows that the power grasps and precision grasps are mostly separate from each other, indicating a high energy barrier between the two. One interpretation is that there is no smooth transition between a power grasp and a precision grasp without a non-force-closure grasp along the transition. Intermediate grasps are scattered around. Nodes that are not colored are grasp types not listed in any existing grasp taxonomy, indicating the manually-defined grasp taxonomy, though carefully collected and designed, may still fall short when facing a large variety of grasps in various applications.

In \cref{fig:tax-b}, we draw various types of power grasps in different colors. Only the power grasps close to the precision grasps belong to the power sphere type. This observation matches our intuition as a power sphere grasp is similar to a precision sphere grasp, with a slight difference in the distance between the object and the palm. In other words, there exists a smooth transition between a precision sphere grasp and a power sphere grasp, such that all snapshots along the transition are force-closure grasps. Please refer to Feix \etal~\cite{feix2015grasp} for details about the power and precision sphere grasps.

\begin{figure}[t!]
    \centering
    \includegraphics[width=\linewidth]{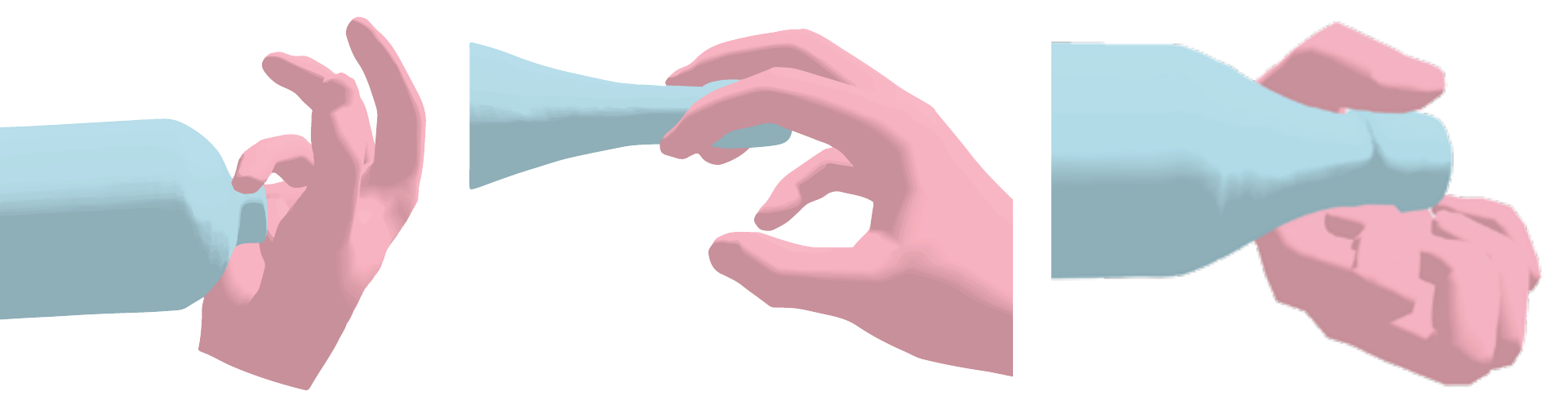}
    \caption{\textbf{Examples of novel grasp poses.} To the best of our knowledge, these newly discovered grasp poses do not correspond to any grasp types in existing human-designed grasp taxonomy (\eg, \cite{taheri2020grab,corona2020ganhand}).}
    \label{fig:rare}
\end{figure}

In \cref{fig:tax-c}, we observe that sphere grasps and tri- or quad-pod grasps are close to each other. This observation is also expected since many sphere grasps can be converted to tri- or quad-pod grasps by merely lifting one or two fingers.

We further demonstrate that our algorithm can find natural but novel grasps in \cref{fig:rare}. These grasps are rarely collected in any of the modern 3D grasp datasets (\eg, \cite{taheri2020grab,corona2020ganhand}), since they do not belong to any type as defined in the grasp taxonomy. However, these grasps are valid grasps and could well exist during physical manipulations. For example, the left example in \cref{fig:rare} is commonly used to twist-open a bottle when some of the fingers are occupied or injured. The second example would occur if one is already holding something (\eg, a ball) in the palm while picking up another bottle-like object.

These grasps occur because the human hand is excellent in doing multiple tasks simultaneously, which have not been recognized or explored in grasp literature as we always assumed otherwise. Such limitation would hinder a robotic hand's capacity from developing to its full potential. Our method paves the way to explore grasp types beyond the grasp taxonomy, which is a crucial step toward exploiting the total capacity of a complex hand structure such as human hands. 

\begin{figure*}[t!]
    \centering
    \hfill
    \begin{subfigure}{0.08\linewidth}
        \includegraphics[width=\linewidth]{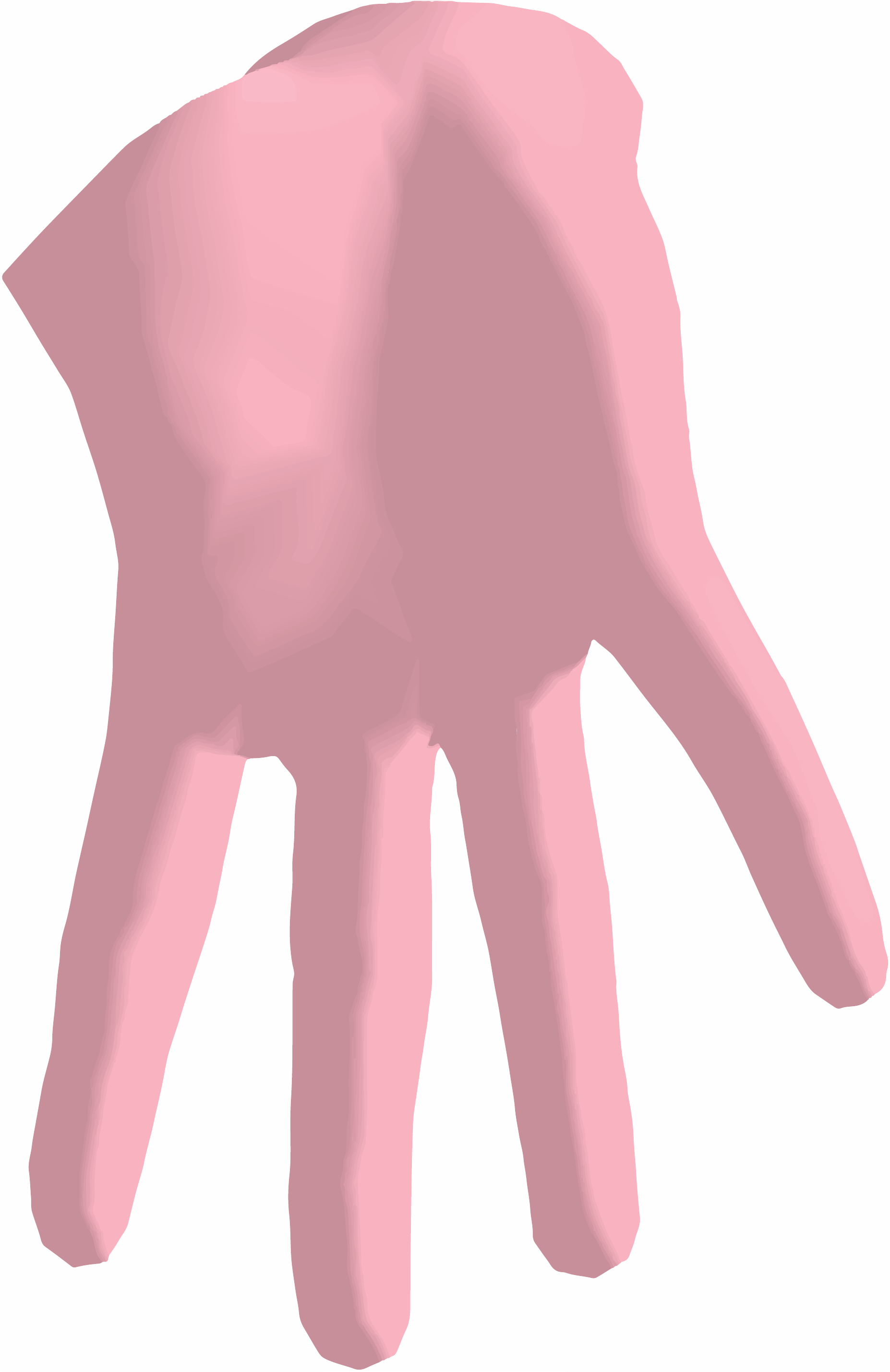}
    \end{subfigure}\hfill
    \begin{subfigure}{0.08\linewidth}
        \includegraphics[width=\linewidth]{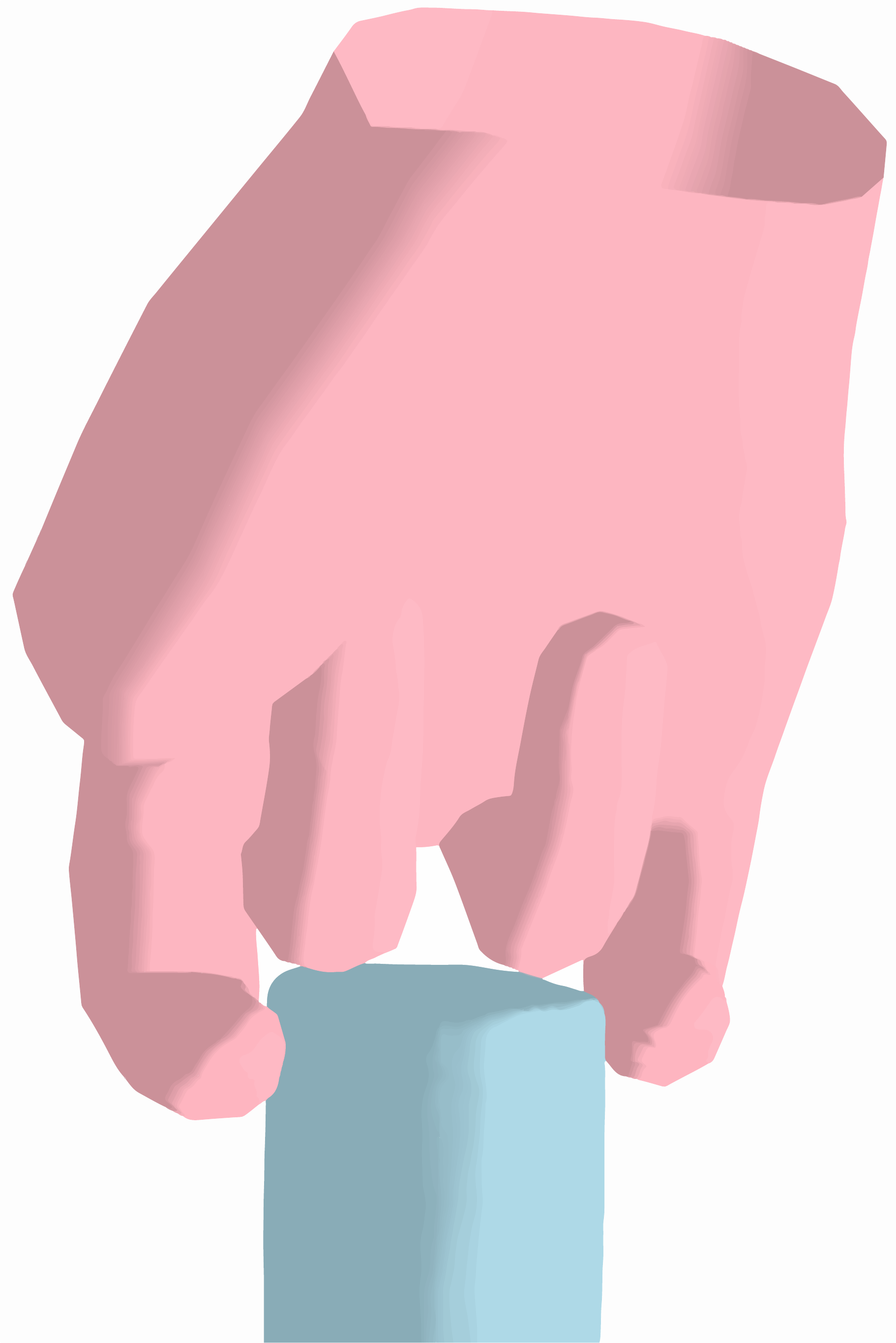}
    \end{subfigure}\hfill
    \begin{subfigure}{0.08\linewidth}
        \includegraphics[width=\linewidth]{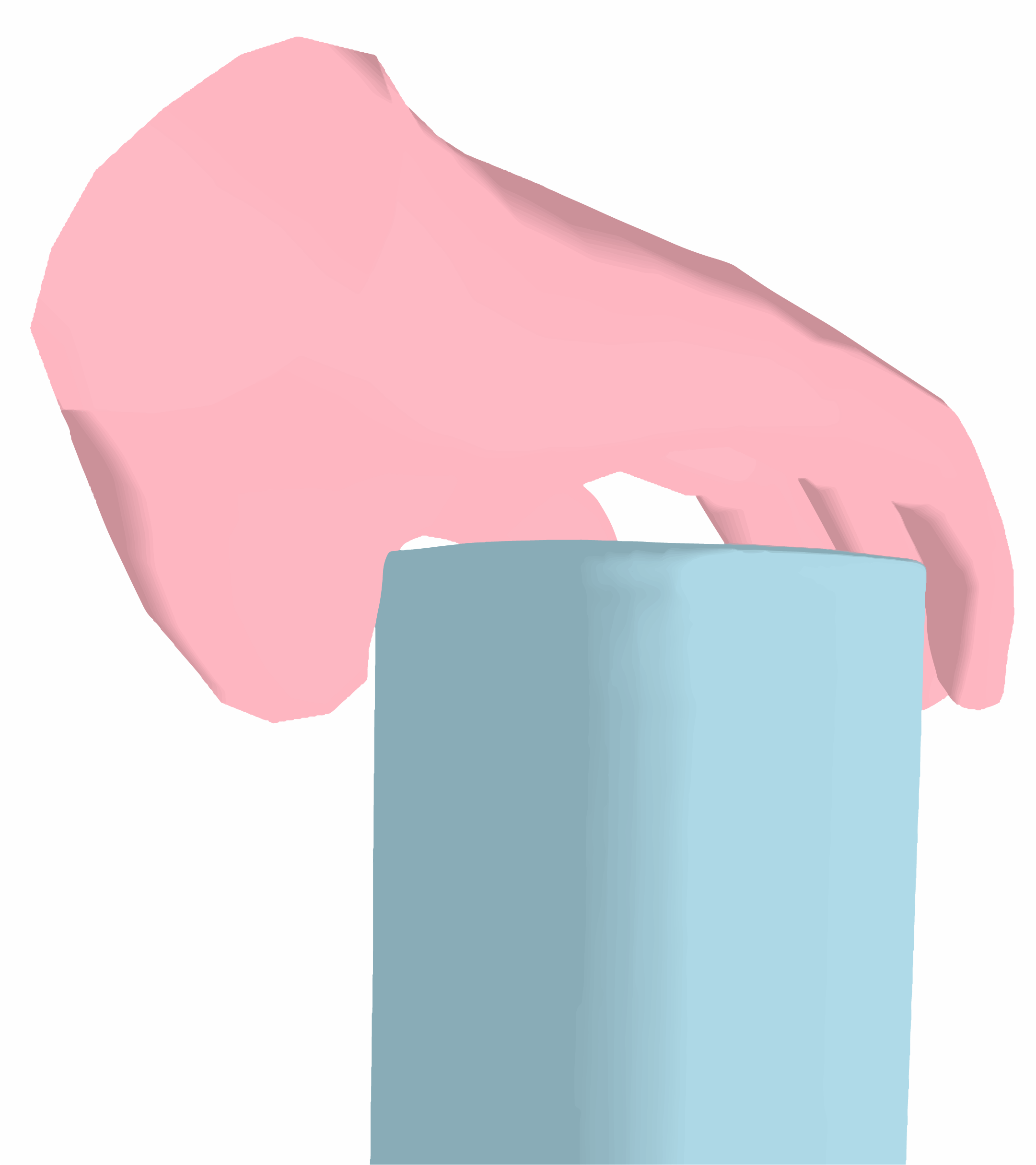}
    \end{subfigure}\hfill
    \begin{subfigure}{0.08\linewidth}
        \includegraphics[width=\linewidth]{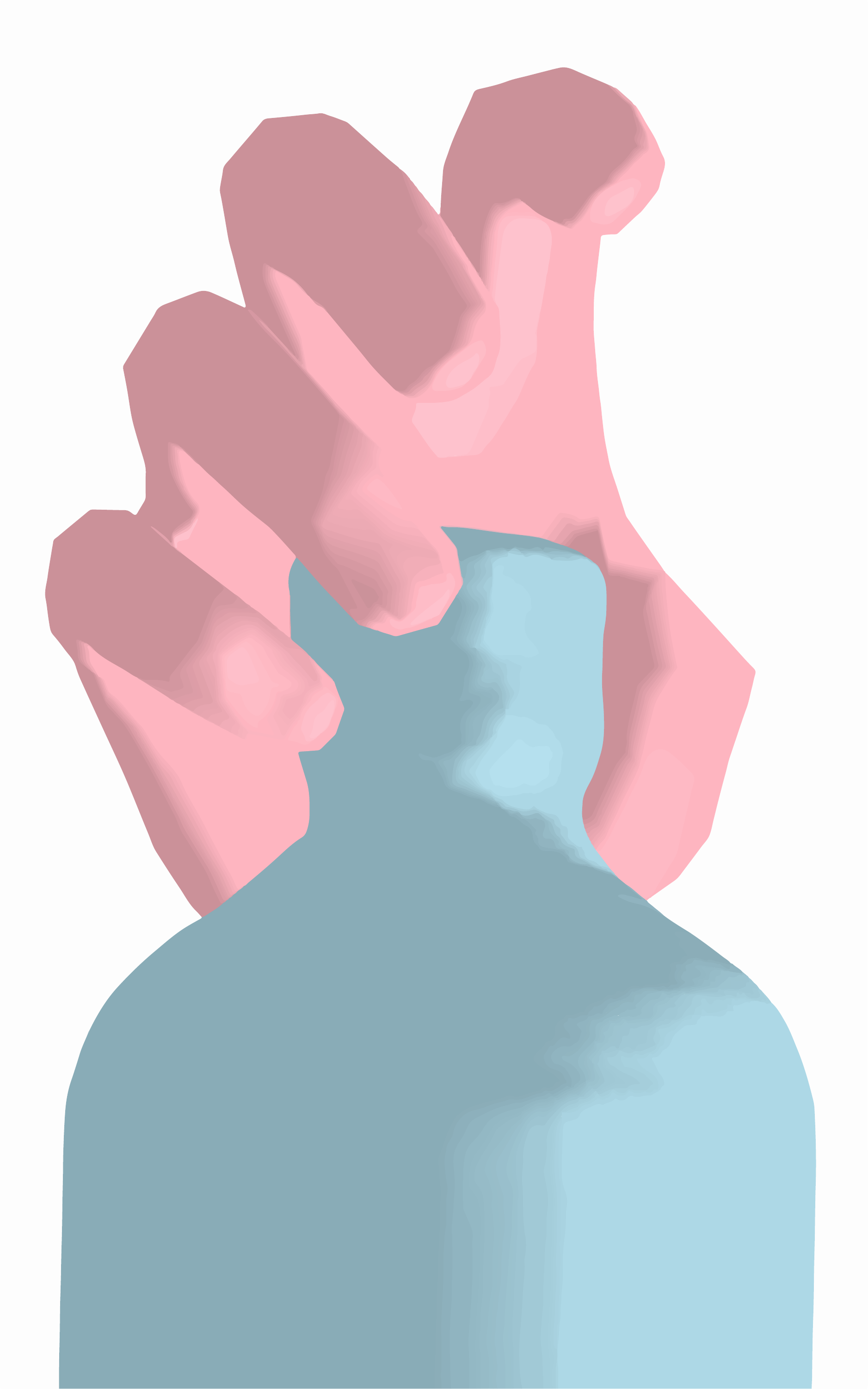}
    \end{subfigure}\hfill
    \begin{subfigure}{0.08\linewidth}
        \includegraphics[width=\linewidth]{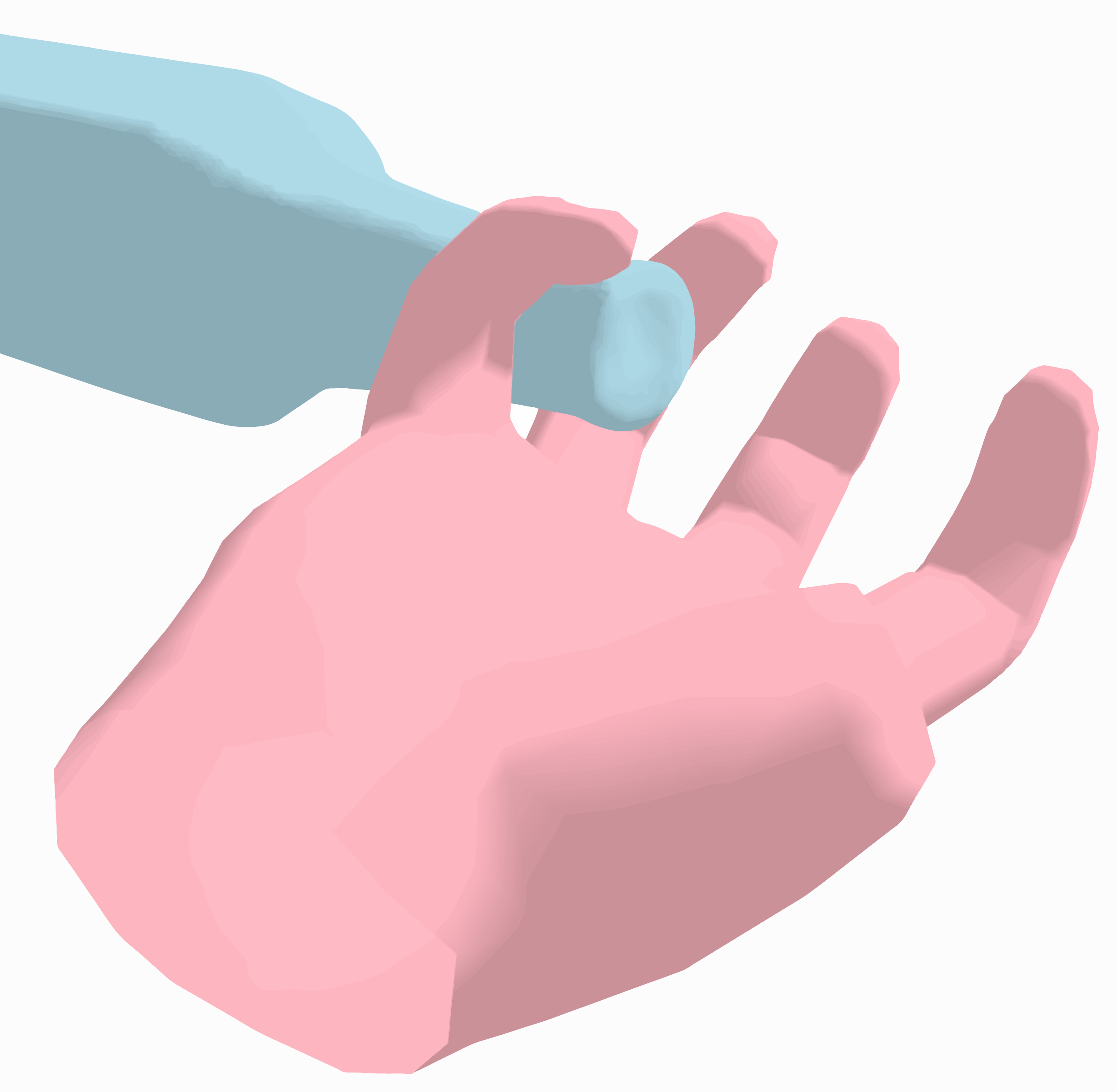}
    \end{subfigure}\hfill
    \begin{subfigure}{0.08\linewidth}
        \includegraphics[width=\linewidth]{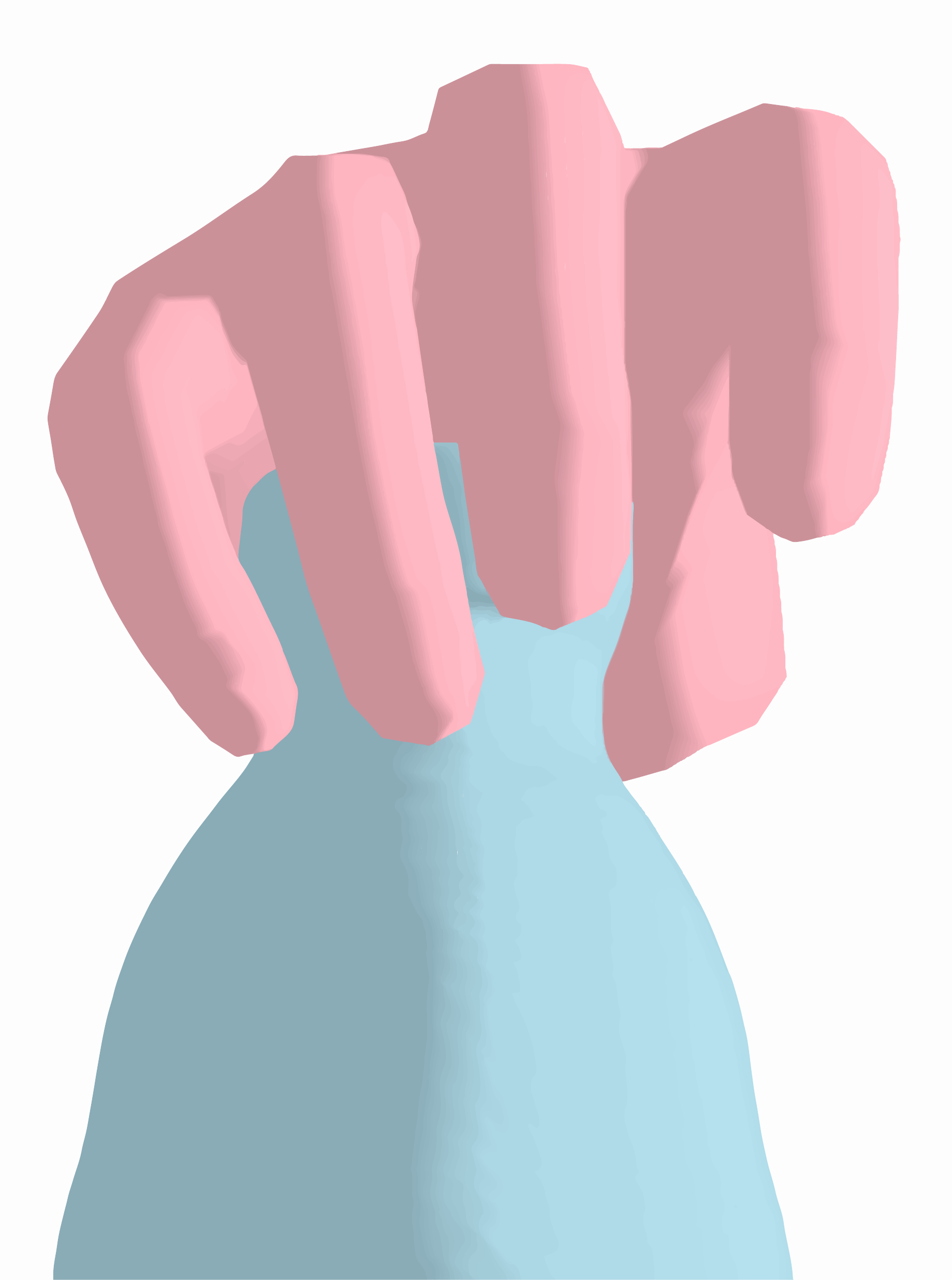}
    \end{subfigure}\hfill
    \begin{subfigure}{0.08\linewidth}
        \includegraphics[width=\linewidth]{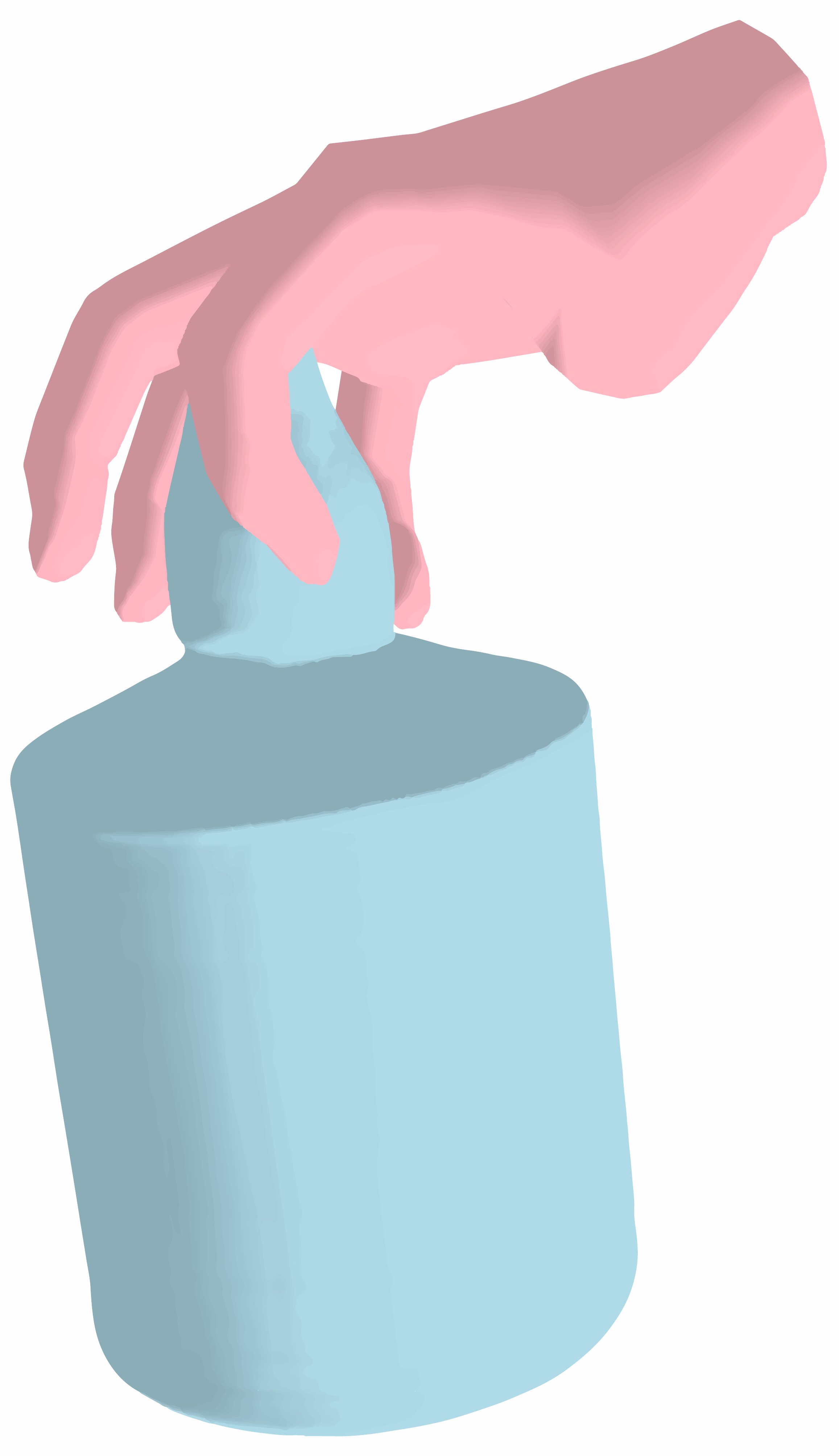}
    \end{subfigure}\hfill
    \begin{subfigure}{0.08\linewidth}
        \includegraphics[width=\linewidth]{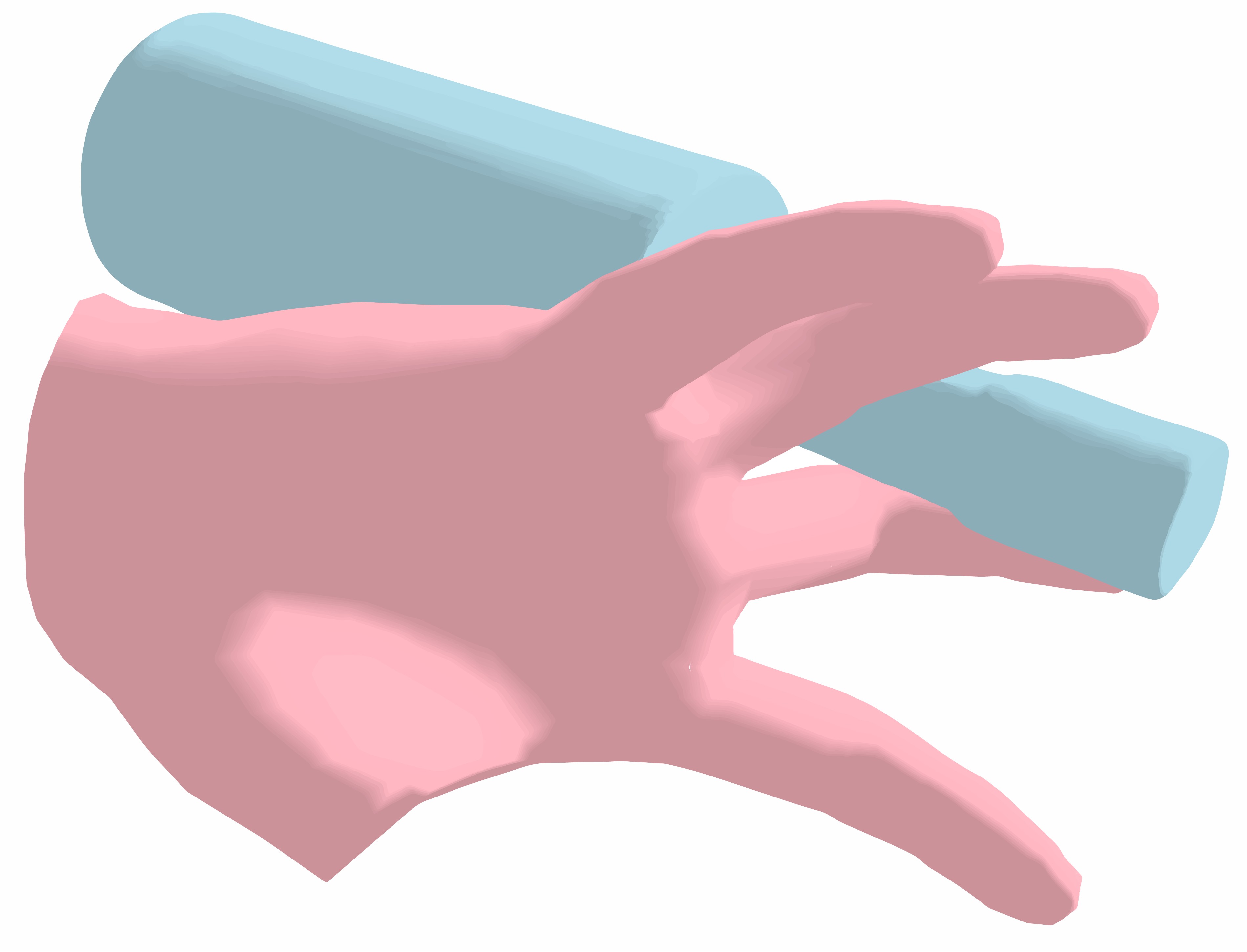}
    \end{subfigure}\hfill
    \begin{subfigure}{0.08\linewidth}
        \includegraphics[width=\linewidth]{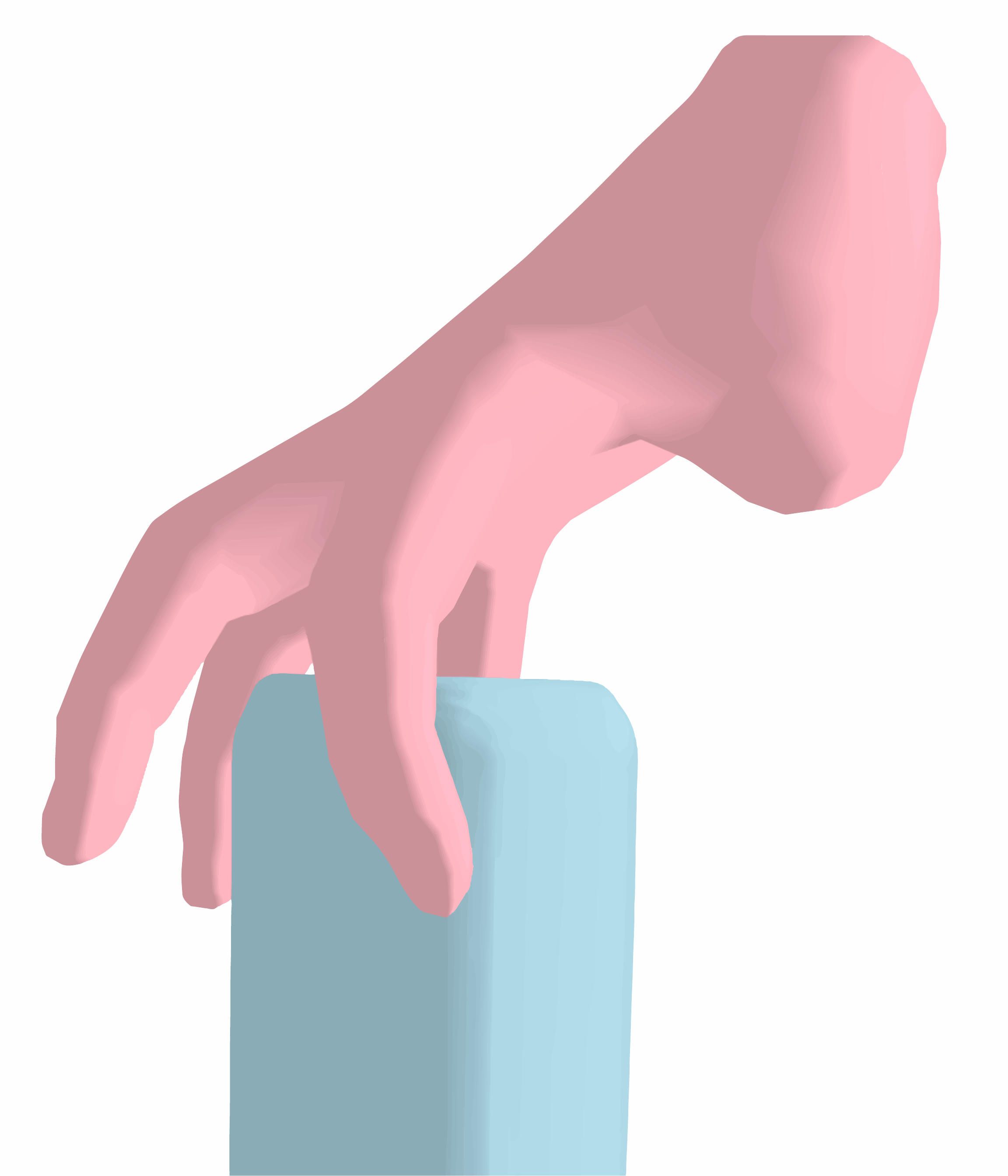}
    \end{subfigure}
    \hfill
    \\
    \hfill
    \begin{subfigure}{0.11\linewidth}
        \includegraphics[width=\linewidth]{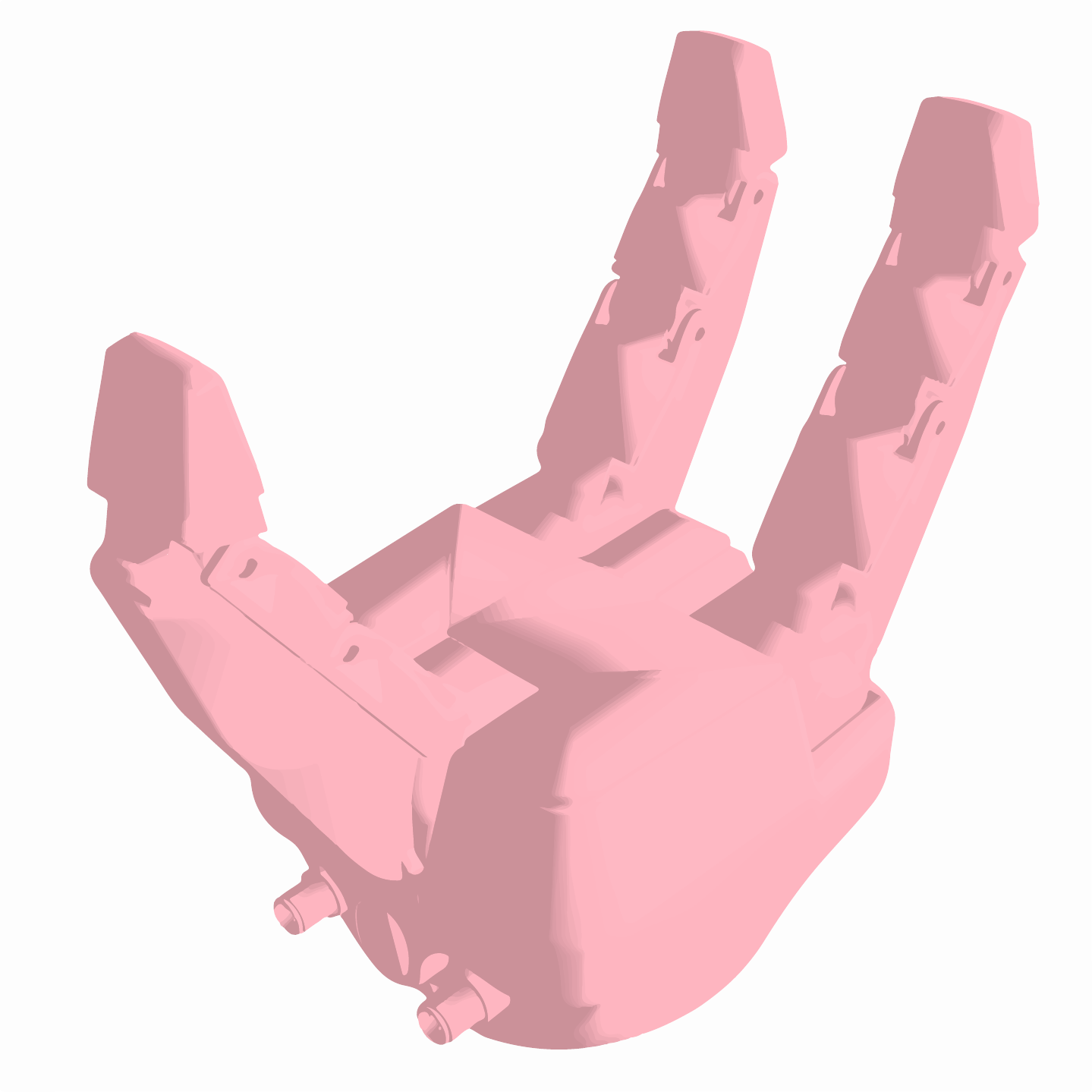}
    \end{subfigure}\hfill
    \begin{subfigure}{0.11\linewidth}
        \includegraphics[width=\linewidth]{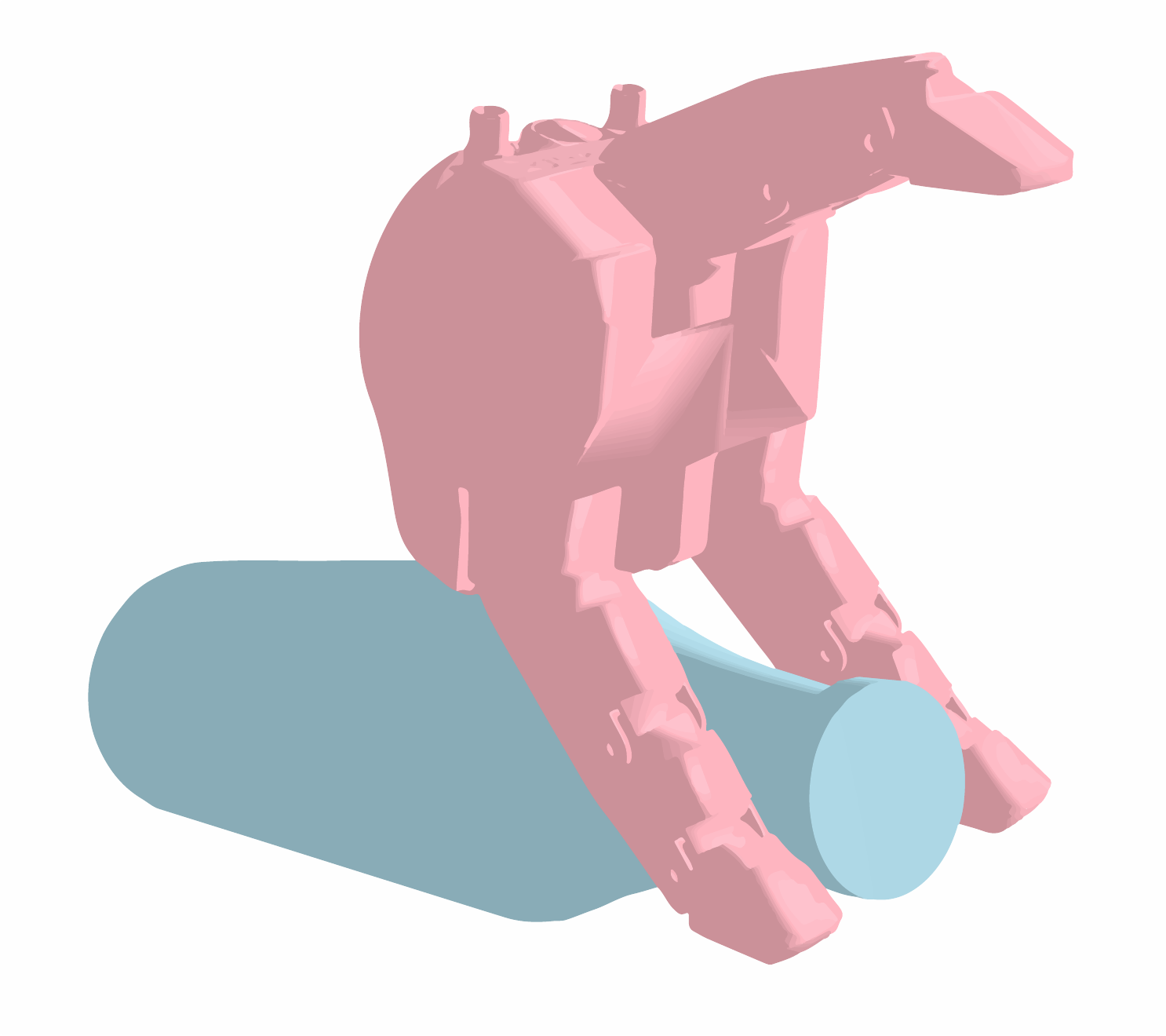}
    \end{subfigure}\hfill
    \begin{subfigure}{0.11\linewidth}
        \includegraphics[width=\linewidth]{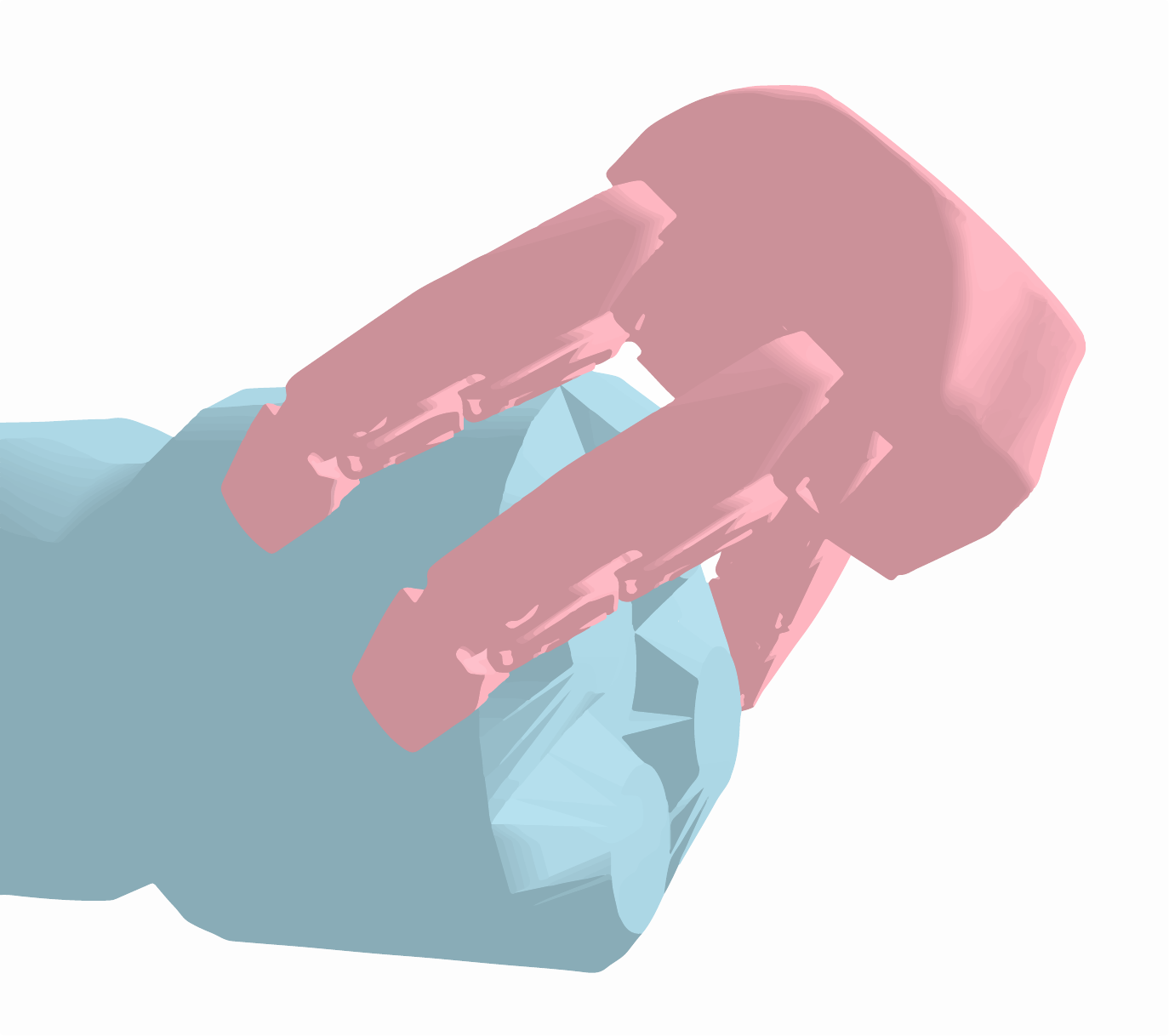}
    \end{subfigure}\hfill
    \begin{subfigure}{0.11\linewidth}
        \includegraphics[width=\linewidth]{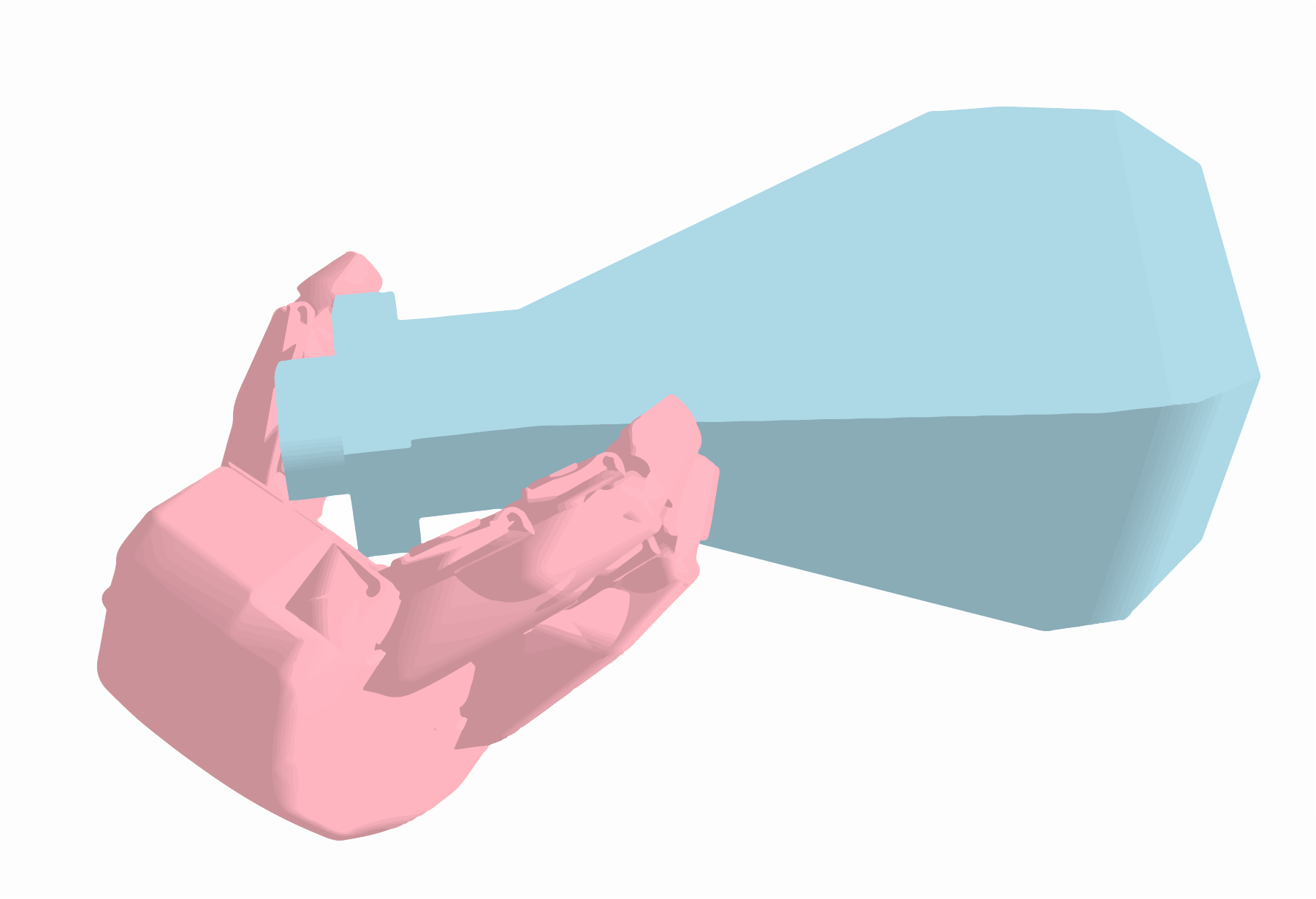}
    \end{subfigure}\hfill
    \begin{subfigure}{0.11\linewidth}
        \includegraphics[width=\linewidth]{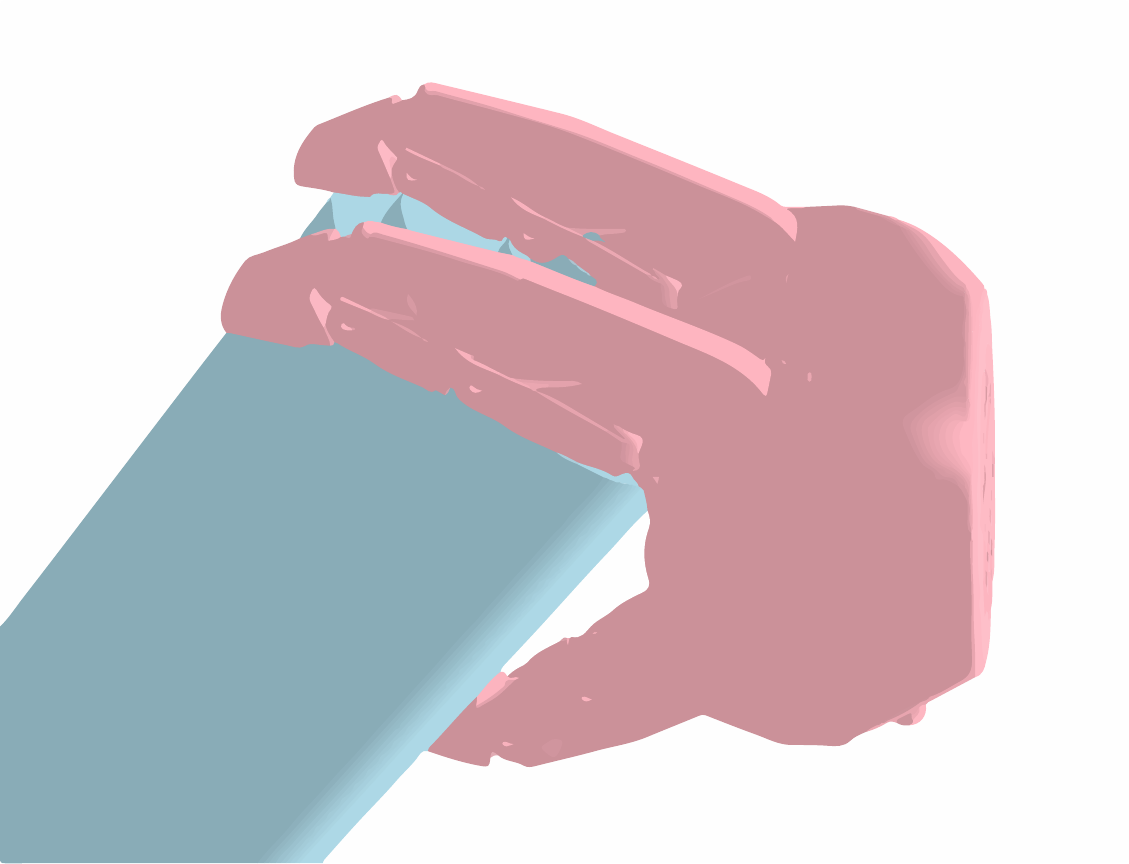}
    \end{subfigure}\hfill
    \begin{subfigure}{0.11\linewidth}
        \includegraphics[width=\linewidth]{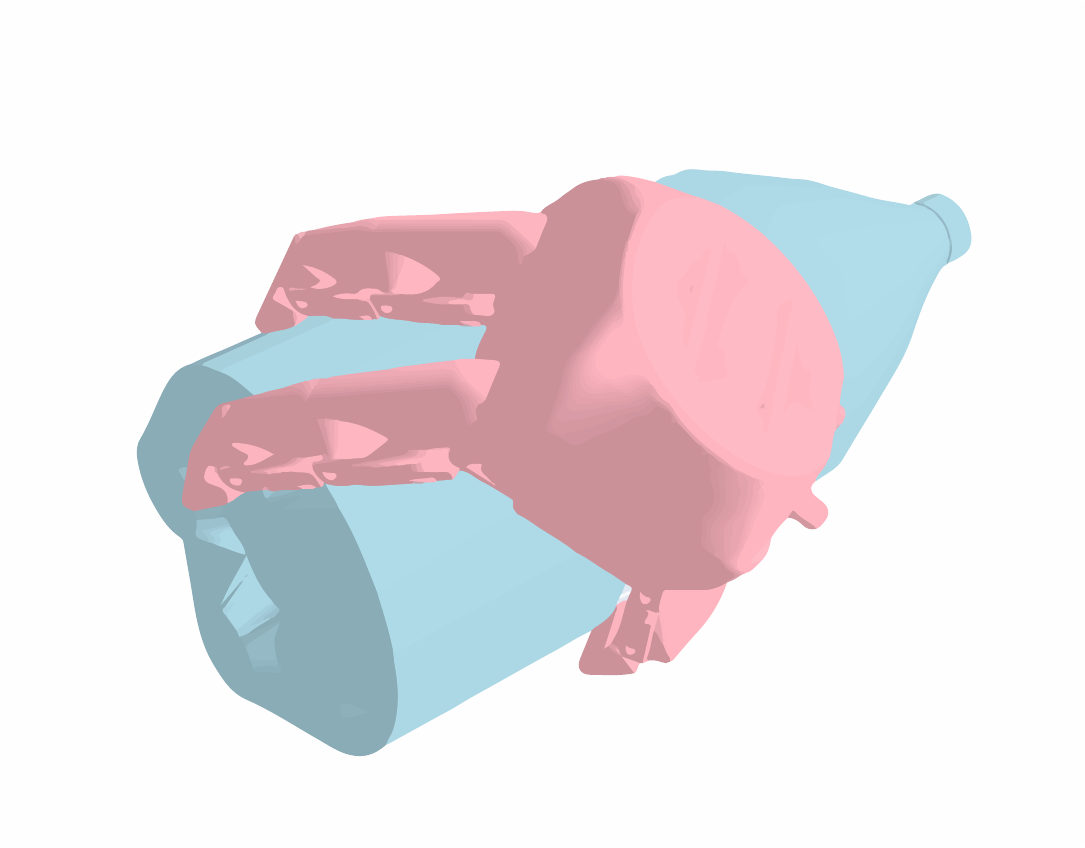}
    \end{subfigure}\hfill
    \begin{subfigure}{0.11\linewidth}
        \includegraphics[width=\linewidth]{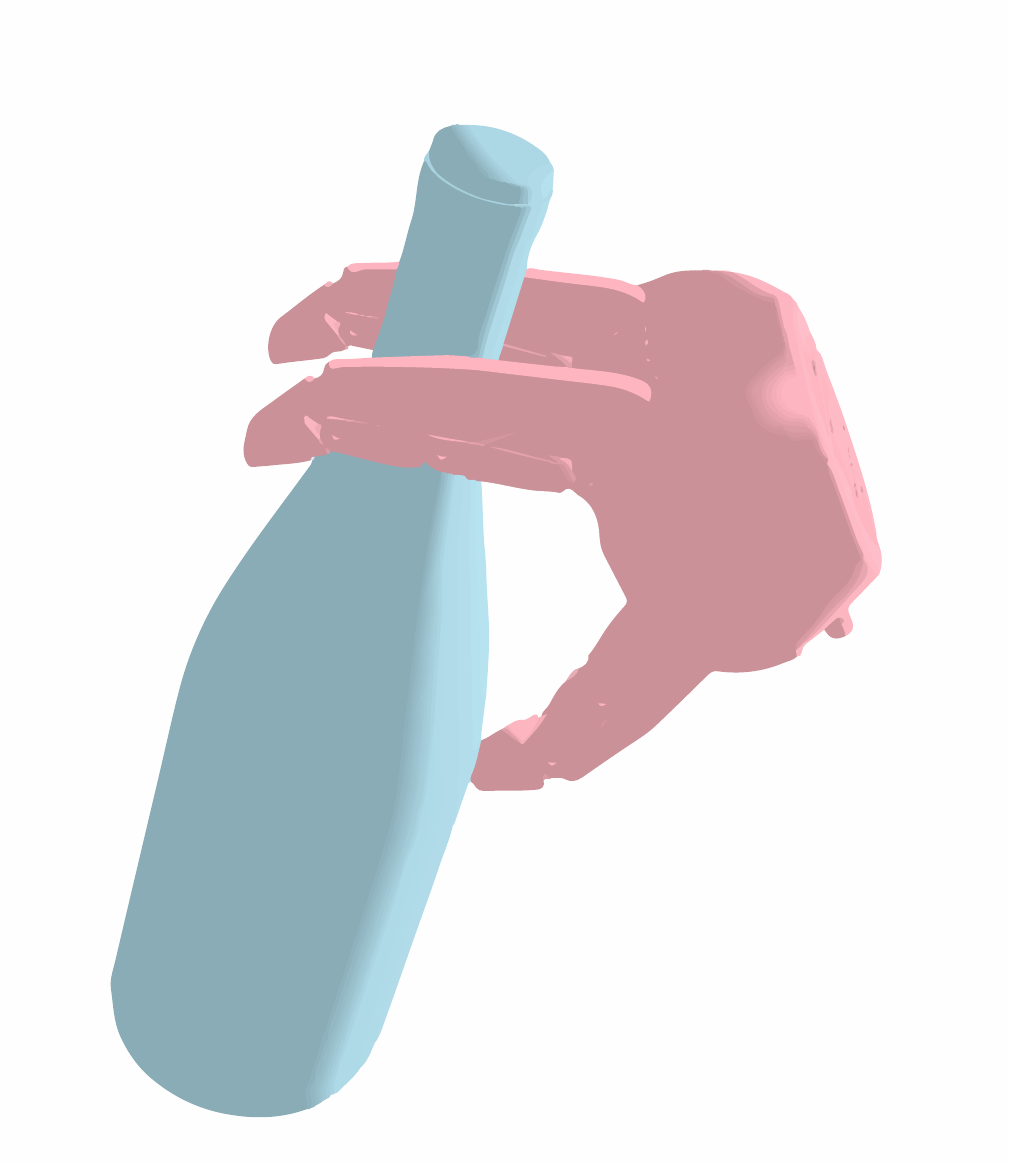}
    \end{subfigure}\hfill
    \begin{subfigure}{0.11\linewidth}
        \includegraphics[width=\linewidth]{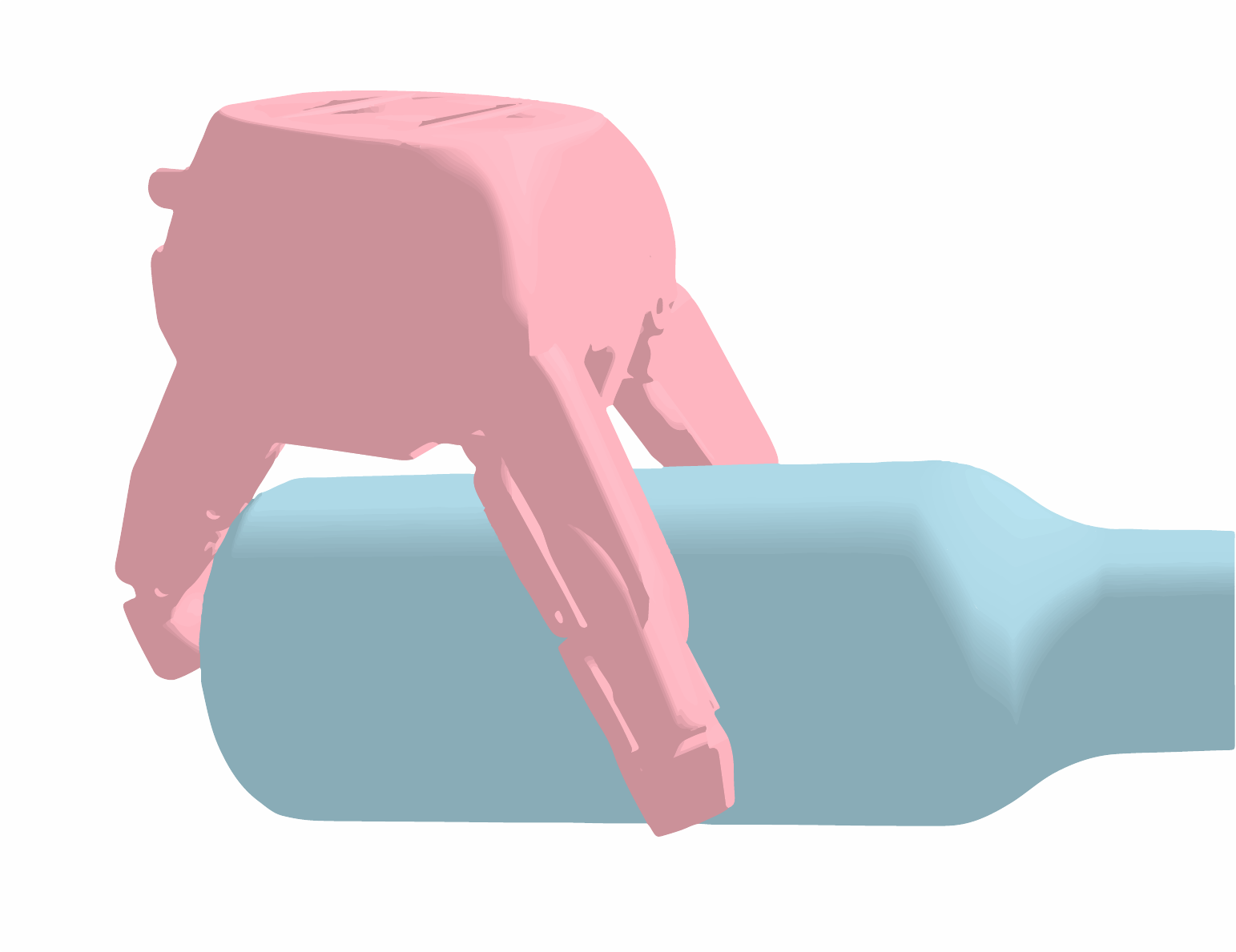}
    \end{subfigure}\hfill
    \begin{subfigure}{0.11\linewidth}
        \includegraphics[width=\linewidth]{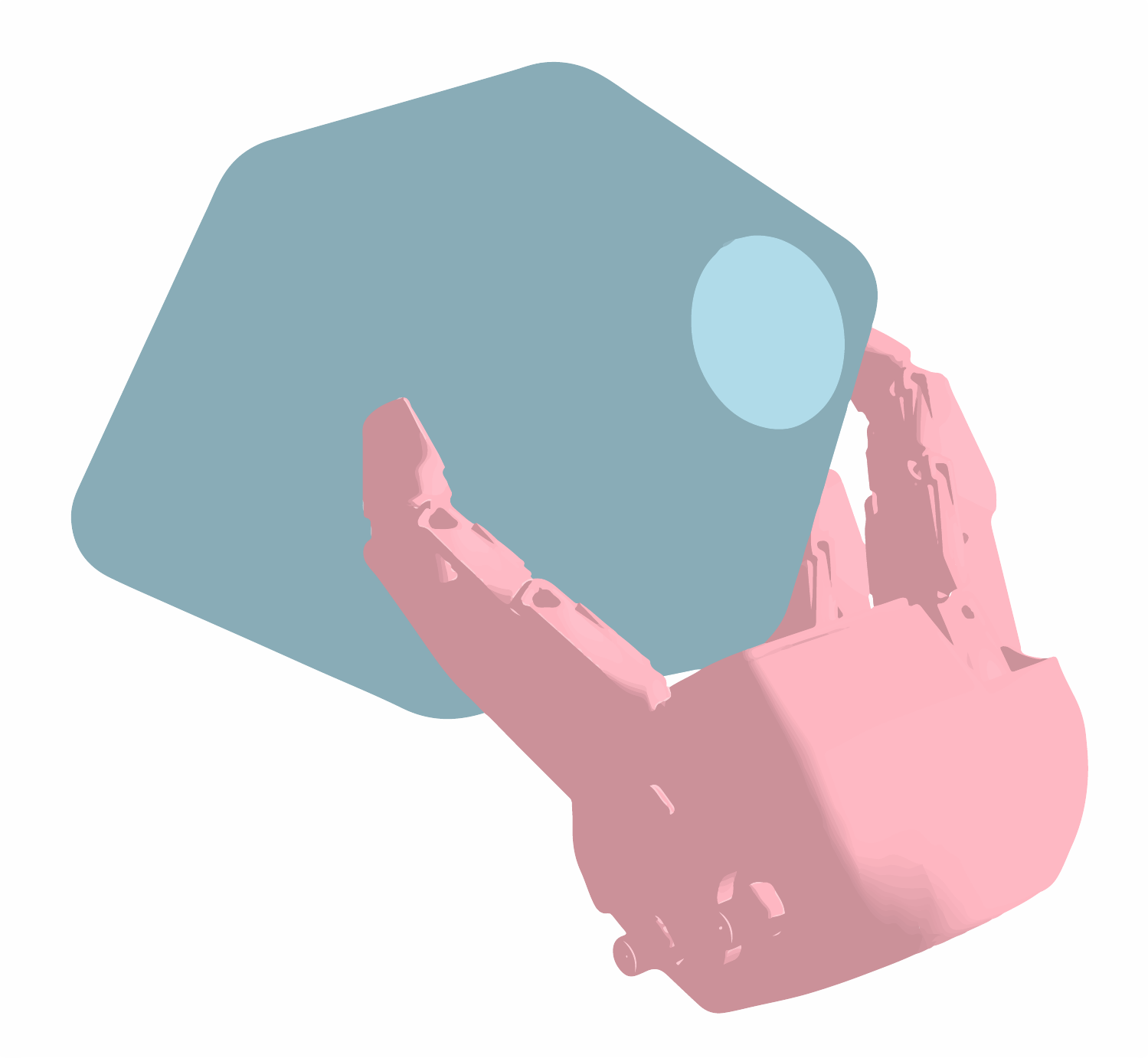}
    \end{subfigure}
    \hfill
    \caption{\textbf{Synthesized grasps of different hands using our formulation.} Top: A MANO hand with its thumb removed. Bottom: A Robotiq 3-finger gripper. The left-most figure shows the hand used in each row.}
    \label{fig:arbitrary}
\end{figure*}

\subsection{Grasp Synthesis for Arbitrary Hand Structures}

Although above experiments primarily rely on MANO for hand modeling and grasp taxonomy, our method in fact makes no assumption on the hand kinematics except for having a differentiable mapping between pose and shape. As a result, we can synthesize grasps for arbitrary hand so long as there exists such a mapping. In \cref{fig:arbitrary}, our method, without modifications, can directly synthesize grasps of a MANO hand with its thumb removed and a Robotiq 3-finger gripper. Specifically, for the 3-finger gripper, we used a differentiable forward kinematics~\cite{sutanto2020encoding} as the mapping from joint states to the hand shape. These examples demonstrate that our method can explore a wide range of grasps for arbitrary hand structure, which could provide valuable insights for understanding the task affordance of prosthetic or robotic hands, and hands with injuries or disabilities. Our method is also applicable to animations for grasps of non-standard hands or claws. 

\begin{figure}
    \centering
    \begin{subfigure}[b]{0.2\linewidth}
        \includegraphics[width=\linewidth]{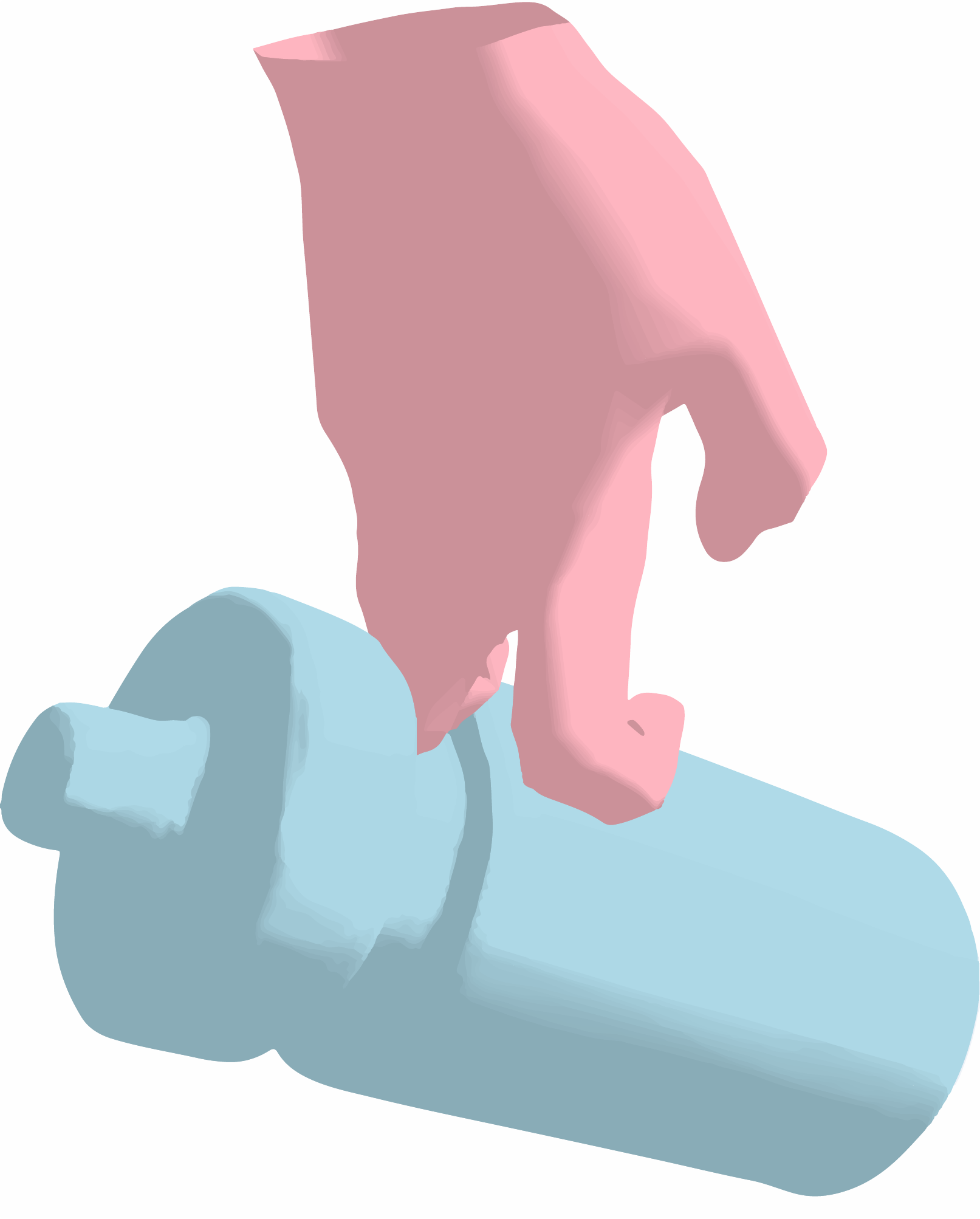}
        \caption{}
    \end{subfigure}
    \begin{subfigure}[b]{0.3\linewidth}
        \includegraphics[width=\linewidth]{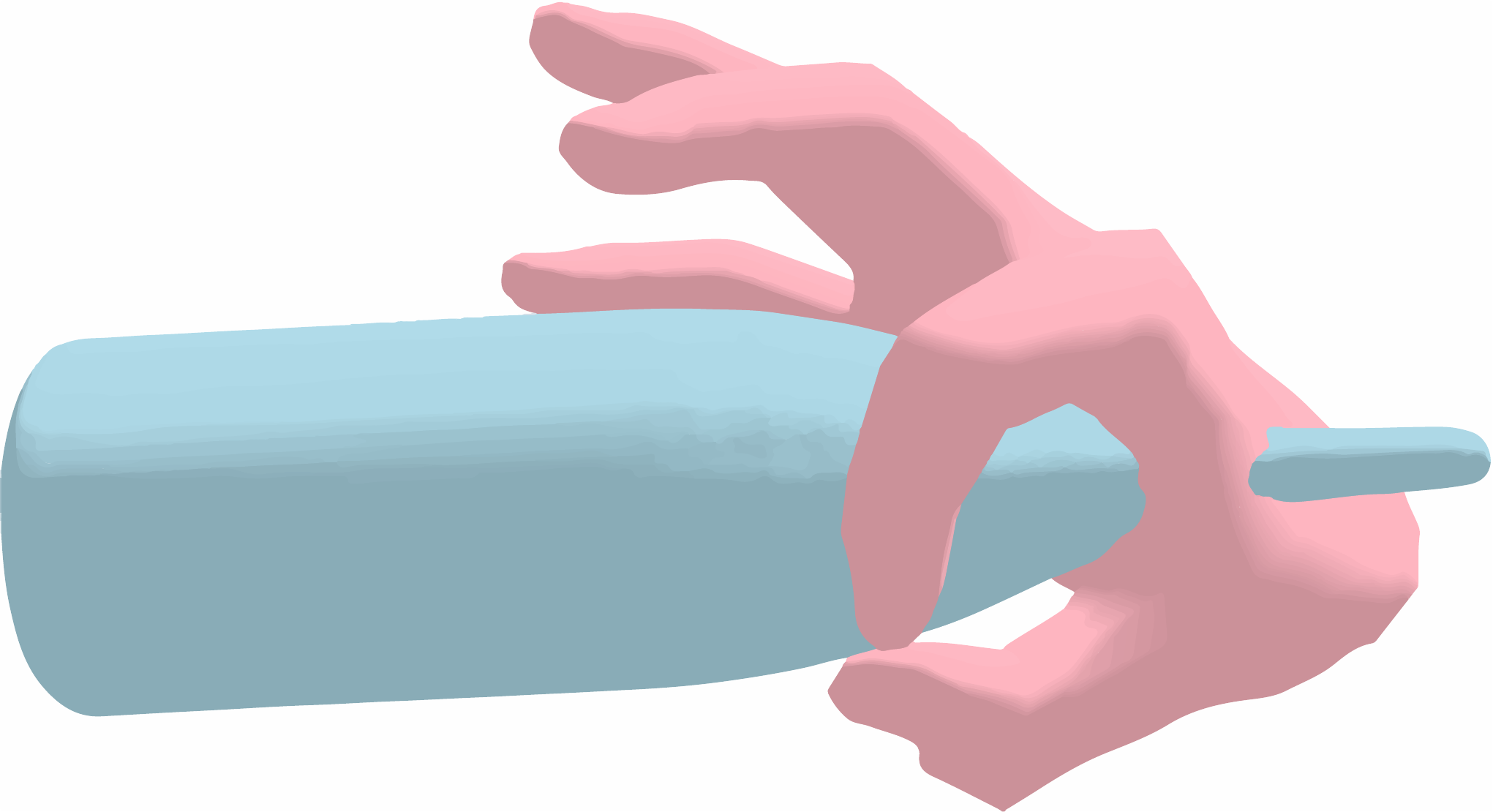}
        \caption{}
    \end{subfigure}
    \caption{\textbf{Examples of failure cases.} (a) Concave shape results in force closure configuration that is not a grasp. (b) Sparse penetration detection leads to intersection.}
    \label{fig:failure}
\end{figure}

\subsection{Limitations}

We show two representative failure cases in \cref{fig:failure}, wherein an unstable or unrealistic grasp receives a low force closure score. In our experiment, most failure cases are caused by concavities in object shapes. For concave shapes, the force closure requirement is sometimes satisfied with a single finger in the concavity, providing contact forces in opposing directions. The issue may be eliminated with manually defined heuristics, such as enforcing contact points on different fingers or encouraging contact points to have larger distances between each other. Another common failure comes from model intersections. We test penetration by computing the signed distance between hand surface vertices and the object shape. When the vertices are sparse, or the object has a pointy part, it is possible for the object to penetrate the hand without being detected. This issue can be addressed with a dense sample of hand surface vertices or by adopting a differentiable mesh intersection algorithm. 

Another primary limitation of our approach lies in the gap between the simulation and the reality. Our algorithm assumes perfect knowledge of the object shape and its signed distance field. Inferring such properties from perception is non-trivial. We plan to address this issue in future studies.

\section{Conclusion}

We formulated a fast and differentiable approximation of the force closure test computed within milliseconds, which enables a new grasp synthesis algorithm. In a series of experiments, we verified that our force closure estimator correctly reflects the quality of a grasp, and demonstrated the proposed grasp synthesis algorithm could generate diverse and physically stable grasps with arbitrary hand structures. The diversity of the generated grasps is validated by its alignment with widely accepted grasp taxonomy with newly discovered grasp types.

We believe that exploring different grasp types is crucial for future work of understanding the hand's total functional capacity, whether it is a prosthetic hand, a robotic hand, or an animated character's hand.

\bibliographystyle{IEEEtran}
\bibliography{reference}
\end{document}